\documentclass[letterpaper]{article} % DO NOT CHANGE THIS
\usepackage[]{aaai23}  % DO NOT CHANGE THIS
\usepackage{times}  % DO NOT CHANGE THIS
\usepackage{helvet}  % DO NOT CHANGE THIS
\usepackage{courier}  % DO NOT CHANGE THIS
\usepackage[hyphens]{url}  % DO NOT CHANGE THIS
\usepackage{graphicx} % DO NOT CHANGE THIS
\usepackage{natbib}  % DO NOT CHANGE THIS AND DO NOT ADD ANY OPTIONS TO IT
\usepackage{caption} % DO NOT CHANGE THIS AND DO NOT ADD ANY OPTIONS TO IT
\usepackage{algorithm}
\usepackage{algorithmic}
\usepackage{amsmath,amssymb,amsfonts}
\usepackage{textcomp}

\usepackage{booktabs, multirow}
\usepackage{hyperref}

\usepackage{newfloat}
\usepackage{listings}
\DeclareCaptionStyle{ruled}{labelfont=normalfont,labelsep=colon,strut=off} % DO NOT CHANGE THIS
\floatstyle{ruled}
\newfloat{listing}{tb}{lst}{}
\floatname{listing}{Listing}
\title{An Empirical Analysis of Fairness Notions under Differential Privacy\thanks{Accepted for a Spotlight Presentation at The Fourth AAAI Workshop on Privacy-Preserving Artificial Intelligence (PPAI-23).} }
\author {
    Anderson Santana de Oliveira,\textsuperscript{\rm 1}
    Caelin Kaplan, \textsuperscript{\rm 2}
    Khawla Mallat \textsuperscript{\rm 1}
    Tanmay Chakraborty \textsuperscript{\rm 3}
}
\affiliations {
    \textsuperscript{\rm 1} SAP\\
    \textsuperscript{\rm 2} SAP and INRIA\\
    \textsuperscript{\rm 3} SAP and Eurecom\\
    firstname.lastname@sap.com
}

\begin{document}

\maketitle

\begin{abstract}
Recent works have shown that selecting an optimal model architecture suited to the differential privacy setting is necessary to achieve the best possible utility for a given privacy budget when using DP-SGD. In light of these findings, we empirically analyze how different fairness notions belonging to distinct classes of statistical fairness criteria (independence, separation and sufficiency) are impacted when one selects a model architecture that is suitable for DP-SGD and optimized for utility. Using standard datasets from ML fairness literature and a rigorous experimental protocol, we show that under these conditions, the differences across groups in the relevant fairness metrics (demographic parity, equalized odds and predictive parity) typically decrease or remain negligible compared to a non-private baseline, where the optimal model architecture was also selected to maximize utility. Our findings challenge the understanding that differential privacy will necessarily exacerbate unfairness in deep learning models trained on biased datasets.
\end{abstract}

\setcounter{secnumdepth}{1}

\section{Introduction}
Due to bias frequently present in the data used for machine learning (ML), systems responsible for high-stakes decisions---such as hiring, healthcare, social benefit allocation, and admissions to higher education---often produce unfair outcomes or perpetuate harmful biases against underprivileged groups~\cite{chouldechova2018frontiers,barocas-hardt-narayanan,mehrabi2021survey}. This concern has motivated substantial research efforts aimed at defining fairness notions, developing interventions in ML training to align with those notions, and applying these approaches across various datasets in supervised learning settings ~\cite{feldman2015certifying,hardt2016equality,zafar2017fairness_www,zafar2017fairness_aistats}. However, fairness is not the only consideration (beyond task-specific model utility) that ML models must take into account. They are also subject to membership inference attacks (MIAs), where an adversary attempts to infer whether specific input examples were used to train the model~\cite{shokri2017membership,hilprecht2019monte}. As differential privacy (DP) is one of the primary defenses against privacy attacks, researchers have begun examining its impact on fairness~\cite{bagdasaryan2019differential,xu2020removing,chang2021privacy,noe2022exploring}. These studies have generally shown that the accuracy reduction caused by DP tends to disproportionately affect underprivileged groups.
% % and other aspects of Responsible AI~\cite{song2019privacy,rice2020overfitting,shokri2021privacy,pawelczyk2023privacy}. Regarding fairness specifically, 
% Alternatively, we show that privacy protection does not always have a negative effect when it comes to disparities across groups. 

In parallel to examining DP’s impact on fairness, other research has focused on improving the utility-privacy trade-off inherent in differentially private stochastic gradient descent (DP-SGD). Recent work has demonstrated that model utility for a certain ($\varepsilon, \delta$)-DP privacy level can be increased significantly by selecting model architectures and training hyperparameters tailored to the unique aspects of DP-SGD, rather than simply applying DP-SGD to the best-performing non-private model~\cite{tramer2020differentially,cheng2022dpnas,de2022unlocking,kurakin2022toward}. Despite these findings, prior studies examining the impact of DP on fairness have generally taken the latter approach, applying DP-SGD to non-optimized models. As a result, these analyses have been conducted using private models that perform substantially worse than their non-private baselines, potentially leading to misleading conclusions. Several other concerns with existing analyses include: (1) the exclusive use of utility-based metrics to evaluate disparate impact, without considering established fairness notions; (2) the exclusion of tabular datasets derived from real-world situations, which are typically studied in ML fairness literature; (3) the removal of small minority groups and their intersections from fairness analysis, which is problematic because sensitive attributes, such as sex and race, have been collected at a level of granularity to include them, indicating their presence in real-world settings.

This paper presents the first exploration of DP-SGD's impact on fairness using ML model architectures and hyperparameters specifically optimized for the DP-SGD setting. To identify the private and non-private models with maximum possible utility, an exhaustive grid search is conducted over a broad range of model design and hyperparameter configurations. In addition to utility-based metrics, differences in demographic parity, equalized odds, and predictive parity across groups are measured using k-fold cross-validation on five commonly studied ML fairness datasets, while retaining naturally occurring minority groups in the data. Ultimately, we observe that selecting model and training configurations tailored to DP-SGD typically reduces or maintains negligible differences across groups in relevant fairness metrics, compared to the non-private baseline. These findings suggest that, contrary to previous reports, DP-SGD does not systematically amplify disparities across groups, particularly when fairness is evaluated using established fairness notions and highly-studied tabular ML fairness datasets.

The remainder of this paper is organized as follows. Section~\ref{sec:background} provides the necessary background on DP and fairness notions for understanding our contributions and discusses related work on the interplay between these two domains. Section~\ref{sec:methodology} details the ML fairness datasets, the model architecture and training hyperparameter configurations explored, and the privacy-preserving model selection process. Section~\ref{sec:experiments} presents the experimental results and provides commentary on the findings. Finally, Section~\ref{sec:conclusion} concludes the paper.

\section{Background}\label{sec:background}
\subsection{Differential Privacy}
Differential privacy (DP) is a rigorous mathematical framework for quantifying and limiting the amount of information that can be inferred about any input data point from the output of an algorithm~\cite{dwork2006calibrating,dwork2006our,dwork2014algorithmic}. Its strong privacy guarantees come from ensuring that the presence or absence of any single data point has a limited effect on the distribution of the algorithm's output. More formally, an algorithm $\mathcal{M}$ satisfies $(\varepsilon,\delta)$-differential privacy if for all set of outputs $S \subseteq \text{Range}(\mathcal{M})$ and for all datasets $D$ and $D'$ that differ on a single example:
\begin{equation*}
\Pr[\mathcal{M}(D) \in S] \leq e^\varepsilon \Pr[\mathcal{M}(D') \in S] + \delta,    
\end{equation*}
where $\varepsilon$ represents the privacy budget (smaller values indicating stronger privacy guarantees) and $\delta$ is the small probability of failure (typically $\delta \ll 1 / N$, where $N$ is the size of $D$~\cite{abadi2016deep}). At a high level, the $(\varepsilon, \delta)$-DP guarantees imply that the probability distribution of the model's outputs changes by at most a factor of $e^\varepsilon$ (with a small additional $\delta$ probability of failure) when trained on two separate datasets that differ by a single example. DP has several particularly useful properties. First, the composition of multiple algorithms with DP guarantees is still differentially private, i.e., if $\mathcal{M}$ and $\mathcal{M}'$ are $(\varepsilon,\delta)$-DP and $(\varepsilon',\delta')$-DP, respectively, then according to the basic composition theorem, $\mathcal{M} \circ \mathcal{M}'$ has in the worst case ($\varepsilon$~+~$\varepsilon'$, $\delta$~+~$\delta'$)-DP guarantees~\cite{dwork2006our, dwork2006calibrating}. Second, the output of any differentially private algorithm can be arbitrarily post-processed without any impact on its privacy guarantees. While other variants of DP exist that use slightly modified formulations (e.g., R{\'e}nyi Differential Privacy~\cite{mironov2017renyi}), $(\varepsilon,\delta)$-DP is presented for its simplicity.

\paragraph{DP-SGD.} As differential privacy was originally developed to be a general framework for privacy-preserving data analysis, it has been able to extend beyond its initial applications in private database queries to ML settings through various approaches, the most common of which is Differentially Private Stochastic Gradient Descent (DP-SGD)~\cite{abadi2016deep}. This technique adapts the standard SGD algorithm to provide DP guarantees for the trained model. The key modifications include:
\begin{enumerate}
    \item Per-example gradient clipping to bound the gradient calculation's sensitivity.
    \item Adding calibrated Gaussian noise to the aggregated gradients.
\end{enumerate}
By applying these operations and choosing $\sigma = \sqrt{2\log\frac{1.25}{\delta}} / \epsilon$, according to standard arguments~\cite{dwork2014algorithmic}, each step of DP-SGD is $(\epsilon,\delta)$-DP with respect to the batch. Moreover, for a training data of size $N$, batch size of size $B$, sampling ratio equal to $\alpha = B / N$, and an overall number of training steps $K = T J$, using the moments account~\cite{abadi2016deep} the algorithm has $(O(\epsilon\alpha\sqrt{K}),\delta)$-DP guarantees with respect to the training data for an appropriate choice of noise scale and gradient norm bound. In essence, this procedure guarantees the privacy of the training data (up to the $(\varepsilon,\delta)$-DP privacy budget) by limiting each individual data point's contribution to the model's learned parameters.

\subsection{Fairness Notions}\label{secNotions}
In this section, we define the group fairness notions to be analyzed, which are drawn from three distinct classes of statistical non-discrimination criteria: independence, separation, and sufficiency~\cite{barocas-hardt-narayanan}. These notions pertain to the joint distributions of sensitive attributes, target labels, and predicted labels, and are expressed as conditional independence statements between these variables~\cite{castelnovo2022clarification}. We use demographic parity to represent independence, equalized odds to represent separation, and predictive parity to represent sufficiency, each of which is described in detail below. Given the subjective nature of fairness criteria selection for different use cases, we do not assess their specific relevance to the datasets in our analysis (see~\cite{makhlouf2021applicability} for a discussion on this topic). Instead, we focus on examining the effects of using DP-SGD to train ML models with respect to each criterion.

\paragraph{Demographic Parity.}
A classifier satisfies the demographic parity constraint when each subgroup receives a positive outcome at equal rates. More formally, this can be expressed as:
\begin{align}
P(\hat{Y} = 1 | A = a) = P(\hat{Y} = 1 | A = b ), \nonumber \: \forall a, b \in A \nonumber
\end{align}
where $\hat{Y}$ is the predicted label, and $A$ is the protected attribute indicating subgroup membership~\cite{barocas-hardt-narayanan}. Demographic parity aims to prevent the reinforcement of historical biases and can support underprivileged groups in the short term by enforcing a positive feedback loop. However, while demographic parity focuses on equal outcomes, it overlooks equality of treatment. This can lead to the issue of ``laziness''~\cite{makhlouf2021applicability,carey2022causal}, also known as ``self-fulfilling prophecy''~\cite{dwork2012fairness}, where a model correctly predicts true positives for privileged groups but randomly selects individuals from underprivileged groups until the proportion selected for each subgroup is equal. Additionally, demographic parity can be misapplied in contexts where disparities are legitimate and unrelated to protected attributes~\cite{dwork2012fairness}.

\paragraph{Equalized Odds.}
A classifier satisfies the equalized odds constraint when it has the same false positive rate and false negative rate for all subgroups. More formally, this can be expressed as:
\begin{align}
P(\hat{Y} = 1 | A = a, Y = y) = P(\hat{Y} = 1 | A = b, Y = y) \nonumber \\ 
\forall a, b \in A, y \in \{0,1\} \nonumber
\end{align}
where  $\hat{Y}$ is the predicted label, $A$ is the protected attribute, and $Y$ is the ground truth label~\cite{hardt2016equality}. Unlike demographic parity, equalized odds uses the ground truth label to ensure equality of treatment, thereby eliminating the laziness issue mentioned above. However, equalized odds can be problematic when the target variable is inherently biased (e.g., a university admissions process that has historically discriminated against a particular group), as this can result in the propagation of existing biases~\cite{castelnovo2022clarification}.

\paragraph{Predictive Parity.}
A classifier satisfies the predictive parity constraint when it has the same precision for all subgroups. More formally, this can be expressed as:
\begin{align}
P(Y = 1 | A = a, \hat{Y} = 1) = P( Y = 1 | A = b, \hat{Y} = 1 ) \nonumber \\ 
\forall a, b \in A \nonumber
\end{align}
where  $Y$ is the ground truth label, $A$ is the protected attribute, and $\hat{Y}$ is the predicted label~\cite{chouldechova2017fair}. Contrary to demographic parity and equalized odds, predictive parity evaluates subgroups based on the model's predictions ($\hat{Y}$) rather than the actual outcomes ($Y$). This approach is often more practical, as prediction-based metrics can be assessed immediately after the model makes decisions, while true outcomes can only be observed in retrospect~\cite{castelnovo2022clarification}

As a general point, it is important to recognize that independence, separation, and sufficiency cannot all hold simultaneously \cite[Chapter 3]{barocas-hardt-narayanan}. This means that in datasets like those used in this paper, where the sensitive attributes are correlated with the target label, attempting to enforce more than one fairness criterion may lead to no feasible solution or a trivial model that performs poorly.

\subsection{Related Work}
In this section, we review existing work that addresses fairness concerns in the context of differential privacy.

\citet{bagdasaryan2019differential} investigates the impact of DP-SGD on underrepresented groups using several image and text datasets. The study finds the data from underrepresented subgroups produces larger gradients, causing the clipping operation to disproportionately limit their influence on the learned model. It is also shown that the random noise introduced by DP-SGD has a more significant impact on these groups. However, in some experiments, group membership is confused with class labels, and underrepresented groups are artificially created through under-sampling. The authors do not address how differential privacy impacts specific fairness notions.

\citet{pujol2020fair} examines fair allotment and decision rule problems using differential privacy. The study shows that the noise introduced by strong privacy guarantees leads to disparate decision errors for minority groups. Notably, the empirical analysis employs extremely small privacy budgets (on the order of $10^{-5}$). 

\citet{xu2021removing} proposes a DP-SGD approach that assigns an adaptive gradient clipping norm per group, aiming to achieve uniform accuracy loss across groups (or classes). The experiments do not explore the impact of this approach on other fairness notions.

\citet{tran2021fairness} evaluates excessive risk as a fairness notion in the context of privacy protection using the ensemble method PATE~\cite{papernot2018scalable}. The study analyzes how various model parameters influence the excessive risk difference across groups and demonstrates potential mitigation strategies for disparate effects by employing soft labels.

\citet{jagielski2019differentially} propose optimization procedures for achieving equalized odds under differential privacy. The work establishes theoretical bounds on the trade-offs between accuracy, privacy, and equalized odds, howing that stricter privacy constraints can increase fairness violations. Experimental results are presented to support this claim.

\citet{ganev2022robin} demonstrate the disparate impact of differential privacy on generative models with respect to minority and majority groups. This impact either reduces or increases the gap between demographic groups in the generated synthetic data, affecting both their proportions relative to the real data and the accuracy of downstream classification tasks.

\citet{cheng2022dpnas} showed that model utility (measured by test accuracy) is significantly improved when a targeted search for the best network architecture is conducted during training with DP-SGD. However, it is important to note that the privacy budget values reported in \cite{cheng2022dpnas} do not appear to account for potential privacy leakage during the architecture search process itself, as discussed in \cite{papernot2022hyperparameter}.

\section{Methodology}\label{sec:methodology}

\subsection{Datasets}\label{secDatasets} 
To assess the real-world impact of DP-SGD on fairness, we used tabular datasets with demographic information that are widely used in ML fairness research~\cite{feldman2015certifying,zafar2017fairness_aistats,chouldechova2017fair,zafar2017fairness_www,barocas-hardt-narayanan,xu2020removing,lahoti2020fairness,chang2021privacy,ganev2022robin,wang2022towards,ezzeldin2023fairfed}. These datasets contain various types of biases in the relationships between sensitive features and the target variable.

\paragraph{Adult:} The UCI Adult dataset~\cite{Dua:2019} contains 49,531 income survey records from the 1994 U.S. census. For our classification task, we use the binarized ``income'' feature as the target variable, predicting whether an individual's income exceeds \$50k per year. The dataset also includes features such as sex, race, age, education, occupation, and marital status, among others.

\paragraph{ACS Income:} The ACS Income dataset~\cite{ding2021retiring} is a more recent alternative to the UCI Adult dataset, addressing certain limitations related to employment status and income reporting. It uses data from the years 2014-2018 and contains 1,599,229 data points. The dataset includes the same features as the Adult dataset, along with additional metadata such as geographical information. The task remains to predict whether an individual's income exceeds \$50k per year. For our analysis, we use a 10\% sample of the ACS survey results from California in 2018.

\paragraph{ACS Employment:} The ACS Employment dataset~\cite{ding2021retiring} is derived from the American Community Survey released by the U.S. Census Bureau, with the prediction target being whether an individual was employed during the year of data collection. We use 10\% of the survey results for California in 2018. The dataset includes features such as educational attainment, marital status, citizenship, and parental employment status, among others.

\paragraph{LSAC.} The Law School dataset~\cite{wightman1998lsac} is from the Law School Admissions Council’s National Longitudinal Bar Passage Study, aimed at predicting whether a candidate would pass the bar exam. It includes law school admission records with features such as gender, race, family income, and GPA-derived metrics. The classification target is the binary feature “isPassBar.”

\paragraph{COMPAS.} The COMPAS dataset~\cite{angwin2016machine} for recidivism risk prediction includes records of convicted individuals, along with their criminal history and demographic features such as gender and race. We use the binary outcome of whether the offender was re-arrested as the target variable for classification. Note that a positive outcome (True) represents an undesirable result. Additionally, the majority group (Black Males) has higher positive rates.

In each of these datasets, fairness and privacy are genuine concerns, as the prediction tasks can be applied in critical domains and the features include sensitive attributes. When specifically considering the privacy risks associated with MIAs, the inclusion of an individual in most of these datasets could potentially reveal private information. For example, in the COMPAS dataset, inclusion indicates that an individual has been convicted of a crime, and in the ACS Employment dataset, one could filter the data to only include non-citizens to predict their unemployment, where inclusion in this segment would reveal that the individual is not a citizen.

\subsection{Training}\label{dp-fairness:training}
During training, we retain all protected attributes (i.e., sex and race) in each dataset.\footnote{We initially conducted a second set of experiments where we removed the two sensitive attributes, sex and race, which are used to construct subgroups for evaluating fairness. However, by doing so, we observed a significant decline in the AUC score across all datasets and settings, making it difficult to accurately assess the impact of DP-SGD on fairness metrics.} This approach aligns with the seminal work by~\citet{dwork2012fairness}, which demonstrates that removing protected attributes does not mitigate bias, as these attributes are often correlated with proxy features---this is indeed the case for the five datasets we consider. We also transform all categorical features into one-hot encoded vectors and standardize all numerical features to have zero mean and unit variance for each dataset. 

To conduct our grid search, we first define a template for a feed-forward neural network, consisting of an input block, a variable number of hidden blocks, and a final softmax layer. Each input and hidden block contains a linear layer, normalization, and dropout, as shown in Figure~\ref{figModel} in Appendix~\ref{app:model-architecture}.

Using this model architecture, we perform a 5-fold cross-validation grid search on each dataset to find the best configurations, aiming to maximize overall model utility. The following list presents the configuration options considered in our search:
\begin{itemize}
    \item[$\bullet$] Learning rate: $10^{-5}, 10^{-4}, 10^{-3}, 10^{-2}$
    \item[$\bullet$] Dropout probability: $0.0, 0.1, 0.2$
    \item[$\bullet$] Number of hidden blocks: $1$ to $3$
    \item[$\bullet$] Batch size: $256, 512$
    \item[$\bullet$] Activation functions: Relu, Tanh
    \item[$\bullet$] Optimizers: Adam, SGD
    \item[$\bullet$] Maximum gradient clipping norm (only relevant for DP): $10^{-3}, 10^{-2}, 10^{-1}$
\end{itemize}
Following previous work~\cite{lahoti2020fairness}, we use the AUC score (area under the receiver operating characteristic (ROC) curve) to quantify utility, which is calculated using false positive rates and true positive rates at different decision thresholds (i.e., the value above which a prediction is considered positive in a binary classification). This metric is suitable for fairness-related tasks because it is insensitive to class imbalance. Moreover, optimizing for utility enables us to compare our results to prior works that apply DP-SGD to the best non-private configuration, and to assess the fairness of private models when achieving fairness is not an explicit goal. We perform the hyperparameter search separately for both DP and non-DP settings using the distributed training framework provided by \textit{Ray Tune}\footnote{\url{https://docs.ray.io/en/latest/tune/index.html}}, with DP-SGD applied via the \textit{Opacus}~\cite{opacus} library. All experiments are run using 2 GPUs on an NVIDIA DGX server. 
% To find the best-performing models in all settings, we perform the search for each of the datasets under four different setups: (1) with and without DP; (2) with and without the protected attributes (sex and race). 

The ($\varepsilon, \delta$)-DP budget for a single run of training with DP-SGD is set to $\varepsilon = 5$ and $\delta = 1 / N$, where $N$ is the size of the training data (a smaller $\varepsilon$ indicates stronger privacy protections). To achieve the desired $\varepsilon$ value, the privacy cost at each iteration is calculated~\cite{abadi2016deep}, with training continuing until the privacy budget is exhausted. Using the basic composition theorem of DP (see Section~\ref{sec:background}), the total privacy cost is estimated at $\varepsilon = 20$, as each of the five training splits is used four times for training (and once for validation). Additionally, the randomized hyperparameter search incurs an extra cost of $\varepsilon = 2$~\cite[Theorem 6]{papernot2022hyperparameter}, and training the final model after the search is complete adds another $\varepsilon = 5$. Thus, the total privacy budget for our procedure is conservatively estimated at $\varepsilon = 27$.

The last step in our procedure is to train models for each dataset in the following three settings: (1) Baseline, which corresponds to training with the best-performing non-private model configuration; (2) Baseline + DP, which uses the Baseline configuration and trains the model with DP-SGD---this mirrors most previous research, where DP-SGD is applied to the baseline model with little or no adjustment (e.g., minor changes to the learning rate); and (3) Best DP Model, which corresponds to training with DP-SGD and the best-performing private model configuration. As a final point, while the overall privacy budget for the {\it Best DP Model} is $\varepsilon = 27$, only $\varepsilon = 5$ is used for training the final model, as the remaining $\varepsilon = 22$ was consumed during the grid search. To ensure a fair comparison, the {\it Basline + DP} model, which does not involve a grid search, is trained using the full privacy budget of $\varepsilon = 27$.

% Here, we use a differential privacy budget (epsilon) equivalent to $27$ for each privacy-preserving version. That is, we explicitly set $\epsilon = 27$ for the {\it Basline + DP} models, whereas we set it to $\epsilon = 5.0$ for the {\it Best DP model} settings, and account for the privacy budget spent during the hyperparameter search phase. 

% Next, we trained models for each of the datasets using the best configuration for three settings: {\it Baseline}, without privacy protection; {\it Baseline+DP} which is the baseline model configuration with the addition of differential privacy;  and {\it Best DP Model} ). It is important to remark that the {\it Baseline+DP} case, we did not adjust any other hyperparameters, and used the privacy budget $\epsilon= 27$. Using the default model with differential privacy approximates what has occurred in most previous research, where differential privacy was added to the baseline models with minor adjustments, such as batch size or learning rate.

\subsection{Evaluation}\label{dp-fairness:evaluation}
We evaluate fairness based on the intersection of subgroups formed by all sex and race categories, which are the protected attributes present in each of the five datasets analyzed. This method enables a more accurate understanding of how underrepresented groups are impacted by private classifiers, in contrast to previous works that create artificial minority groups by undersampling certain classes (e.g., reducing the occurrence of class ``8'' in MNIST~\cite{bagdasaryan2019differential}), which has little relevance to real-world scenarios.

As outlined in Section~\ref{sec:background}, we consider three fairness metrics, each corresponding to a statistical fairness criterion: demographic parity (independence), equalized odds (separation), and predictive parity (sufficiency). These metrics are calculated using the \textit{Fairlearn} library~\cite{kulshrestha2021fairlearn}. Following previous work~\cite{lahoti2020fairness}, we focus on utility and fairness metrics for the worst-performing subgroup. Specifically, for the AUC and each fairness metric, we calculate the difference between the best-performing and worst-performing subgroups, which captures the largest disparity in the dataset. The closer these ``difference'' values are to $0$, the fairer the model is with respect to that metric. While the AUC is calculated across various decision thresholds, calculating the fairness metrics requires selecting a specific threshold to obtain a single set of predictions. In this case, we set the decision threshold to $0.5$ for all datasets. All metrics are evaluated on a hold-out test set (i.e., not used during the 5-fold cross-validation), with the results representing the mean values from 10 independent training runs.

\section{Experiments}\label{sec:experiments}
\begin{table*}[ht!]
\centering
\small
\begin{tabular}{c c ccc}
    \toprule
    Dataset & Metric  &  Baseline & Baseline + DP & Best DP Model \\
    \midrule
\multirow{4}{*}{Adult}
      & Overall AUC	 &	 \textbf{0.9056} $\pm$ 0.0011 &	 0.8476 $\pm$ 0.0073 &	 0.9005 $\pm$ 0.0009 \\
	 & AUC Difference	 &	 \textbf{0.1264} $\pm$ 0.0249 &	 0.2226 $\pm$ 0.0556 &	 0.1841 $\pm$ 0.0599 \\
	 & Demographic Parity Difference	 &	 0.2750 $\pm$ 0.0155 &	 0.4683 $\pm$ 0.0612 &	 \textbf{0.2375} $\pm$ 0.0207 \\
	 & Equalized Odds Difference	 &	 \textbf{0.7845} $\pm$ 0.0492 &	 0.8005 $\pm$ 0.0472 &	 0.8000 $\pm$ 0.0000 \\
	 & Predictive Parity Difference	 &	 0.9400 $\pm$ 0.0966 &	 \textbf{0.7567} $\pm$ 0.1933 &	0.8000 $\pm$ 0.000 \\
    \addlinespace
    \hline
    \addlinespace
\multirow{4}{*}{ACS Inc.}
      & Overall AUC	 &	 \textbf{0.8878} $\pm$ 0.0011 &	 0.8155 $\pm$ 0.0045 &	 0.8820 $\pm$ 0.0008 \\
	 & AUC Difference	 &	 0.2546 $\pm$ 0.0569 &	 0.3498 $\pm$ 0.0867 &	 \textbf{0.2225} $\pm$ 0.0219 \\
	 & Demographic Parity Difference	 &	 0.4223 $\pm$ 0.0613 &	 0.6105 $\pm$ 0.1234 &	 \textbf{0.2556} $\pm$ 0.0490 \\
	 & Equalized Odds Difference	 &	 0.4360 $\pm$ 0.0780 &	 0.8499 $\pm$ 0.1656 &	 \textbf{0.3756} $\pm$ 0.0019 \\
	 & Predictive Parity Difference	 &	 \textbf{0.3550} $\pm$ 0.0518 &	 0.7882 $\pm$ 0.2009 &	 0.4032 $\pm$ 0.0271 \\
    \addlinespace
    \hline
    \addlinespace
\multirow{4}{*}{ACS Emp.}  
 & Overall AUC	 &	 \textbf{0.8837} $\pm$ 0.0011 &	 0.8110 $\pm$ 0.0062 &	 0.8702 $\pm$ 0.0013\\
	 & AUC Difference	 &	 0.3401 $\pm$ 0.0875 &	 0.3134 $\pm$ 0.0653 &	 \textbf{0.2073} $\pm$ 0.0480\\
	 & Demographic Parity Difference	 &	 0.4383 $\pm$ 0.0805 &	 0.5317 $\pm$ 0.1561 &	 \textbf{0.3154} $\pm$ 0.0359\\
	 & Equalized Odds Difference	 &	 0.5884 $\pm$ 0.1360 &	 0.6623 $\pm$ 0.1701 &	 \textbf{0.2874} $\pm$ 0.0534\\
	 & Predictive Parity Difference	 &	 0.5534 $\pm$ 0.0998 &	 0.4833 $\pm$ 0.1296 &	 \textbf{0.2968} $\pm$ 0.0674\\
    \addlinespace
    \hline
    \addlinespace
\multirow{4}{*}{LSAC}
	 & Overall AUC	 &	 \textbf{0.8343} $\pm$ 0.0029 &	 0.7755 $\pm$ 0.0125 &	 0.7962 $\pm$ 0.0077 \\
	 & AUC Difference	 &	 0.0435 $\pm$ 0.0056 &	 \textbf{0.0422} $\pm$ 0.0142 &	 0.0575 $\pm$ 0.0128 \\
	 & Demographic Parity Difference	 &	 0.3064 $\pm$ 0.0653 &	 0.2007 $\pm$ 0.0437 &	 \textbf{0.1687} $\pm$ 0.0151 \\
	 & Equalized Odds Difference	 &	 0.2548 $\pm$ 0.0862 &	 0.2853 $\pm$ 0.0467 &	 \textbf{0.1975} $\pm$ 0.0722 \\
	 & Predictive Parity Difference	 &	 \textbf{0.1688} $\pm$ 0.0485 &	 0.2202 $\pm$ 0.0061 &	 0.2197 $\pm$ 0.0082 \\
    \addlinespace
    \hline
    \addlinespace
\multirow{4}{*}{Compas}
      & Overall AUC	 &	 \textbf{0.6895} $\pm$ 0.0041 &	 0.5349 $\pm$ 0.0359 &	 0.6863 $\pm$ 0.0030 \\
	 & AUC Difference	 &	 0.1162 $\pm$ 0.0273 &	 0.1295 $\pm$ 0.0390 &	 \textbf{0.0824} $\pm$ 0.0264 \\
	 & Demographic Parity Difference	 &	 0.5101 $\pm$ 0.0209 &	 \textbf{0.3322} $\pm$ 0.1331 &	 0.3694 $\pm$ 0.0230 \\
	 & Equalized Odds Difference	 &	 0.5592 $\pm$ 0.0476 &	 0.3905 $\pm$ 0.1356 &	 \textbf{0.3726} $\pm$ 0.0375 \\
	 & Predictive Parity Difference	 &	 0.3347 $\pm$ 0.0749 &	 \textbf{0.3009} $\pm$ 0.1052 &	 0.3168 $\pm$ 0.0467 \\
    \addlinespace  
    \bottomrule
\end{tabular}
\caption{Mean and standard deviation values are reported for 10 training runs of the best model configurations. The difference metrics represent the largest disparity for a given metric between any two subgroups in the dataset. The total privacy budget is $\varepsilon = 27$ for both the {\it Baseline + DP} and {\it Best DP} models.\label{tabMetrics}}
\end{table*}

Table~\ref{tabMetrics} shows our results on the five datasets for each of the relevant metrics: Overall AUC, Maximum AUC Difference, Maximum Demographic Parity Difference, Maximum Equalized Odds Difference, and Maximum Predictive Parity Difference. As explained in Section~\ref{dp-fairness:evaluation}, lower values for all ``difference'' metrics indicate better performance. The large values for all the differences in the \textit{Baseline} model column confirm the presence of biases in the datasets. While there is no universal scale for measuring sufficiency, separation, and independence, the differences across groups are significant enough to suggest that, in most cases, this level of disparity would be unacceptable, requiring fairness interventions to reduce disparities based on the chosen fairness criterion for each dataset. Our goal is to observe how these fairness metrics are impacted by the use of DP-SGD. Group-specific metrics for each dataset are provided in Appendix~\ref{app:group-specific}.

We observe that the {\it Best DP Model} outperforms the {\it Baseline + DP} model across nearly all metrics and datasets. Although we do not perform any additional fine-tuning (e.g., adjusting the learning rate) for the {\it Baseline + DP} models, it is still surprising how poorly they perform, especially considering that they use a much larger privacy budget of $\varepsilon = 27$ for training the final model, compared to $\varepsilon = 5$ for the {\it Best DP Model}. This difference in performance serves as further evidence that finding the best-performing model architecture and hyperparameters is essential in the DP setting. Moreover, the {\it Best DP Model} maintains strong Overall AUC scores compared to the \textit{Baseline} models, with decreases typically within 0.01, and a maximum decrease of 0.0381 on the LSAC dataset. 

In terms of fairness metrics, instead of exacerbating subgroup differences, the {\it Best DP Model} actually reduces bias more often than both the \textit{Baseline} and {\it Baseline + DP} models. The clearest examples of this are seen in the ACS Income and ACS Employment datasets, where the Overall AUC is nearly equal to the non-private baselines, and all group disparities have significantly decreased. In contrast, we do not observe substantial reductions in disparities for the Adult dataset, but training with DP-SGD does not appear to worsen biases. Overall, compared to the \textit{Baseline} model, the {\it Best DP Model} reduces disparities for the AUC Difference on three datasets, for the Demographic Parity Difference on all five datasets, for the Equalized Odds Difference on four datasets, and for the Predictive Parity Difference on three datasets.

While this analysis has been empirical, one can hypothesize that the noise and gradient clipping introduced by DP-SGD may act as a regularization effect, potentially enhancing the private model's generalization capabilities. This hypothesis is supported by group-specific metrics evaluated on the Cartesian product of the sensitive attributes for sex and race (see Appendix~\ref{app:intersection}), where the {\it Best DP Model} often assigns positive outcomes more frequently to underprivileged (or minority) groups. However, a more formal analysis along the lines of~\citet{tran2021differentially} is required before drawing any definitive conclusions.

\section{Conclusion}\label{sec:conclusion}
Previous research has suggested that training deep learning models with differential privacy has a disparate impact on accuracy for underprivileged groups. While this observation may hold true when using DP-SGD with an existing non-private model configuration, our experiments show that this is not always the case. Specifically, identifying a model architecture and training hyperparameters suited to the DP setting appears to mitigate many of the previously observed issues with utility metrics and, in fact, often improves fairness when evaluated using standard fairness measures such as demographic parity, equalized odds, and predictive parity. However, our conclusions are based solely on tabular datasets commonly used in ML fairness research. Future work should explore whether these findings hold for datasets outside the tabular domain, such as those used to train language and vision models. Additionally, a deeper understanding of the mathematical relationship between DP-SGD and these fairness notions could provide greater insight into the effects at play. Overall, this research highlights the potential to balance utility with various Responsible AI principles, demonstrating that both can be effectively addressed in certain contexts.

\bibliography{bib}

\onecolumn
\appendix

\section*{Appendix}

\section{Model Architecture}\label{app:model-architecture}
\begin{figure}[h!]
\center
\includegraphics[scale=0.45]{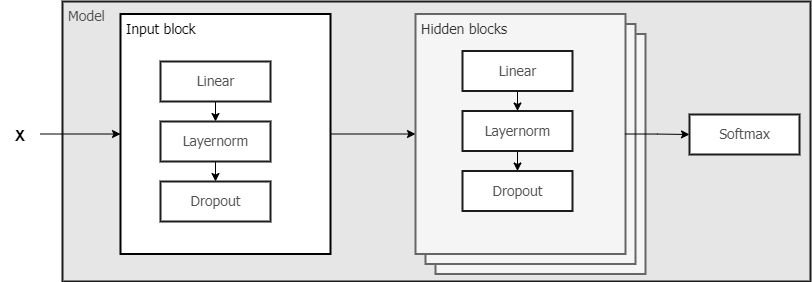}
\caption{Representation of our predefined model architecture. The number of hidden blocks one of the model design choices included in our grid search.} \label{figModel}
\end{figure}

\section{Group distributions in the datasets}\label{app:group-specific}
In this section, the tables display the number of instances in the training data for each group, across all datasets, including both the training and hold-out test data.

\begin{table}[h!]
\centering
\small
\begin{tabular}{c c c c}
Race   & Sex & Train & Test \\ 
\toprule

 Amer-Indian-Eskimo   & Female &    125  &  60\\ 
                      & Male &       214 &  71\\ 
 Asian-Pac-Islander   & Female &    379  & 138\\ 
                      & Male &       703 & 299 \\  
 Black                & Female &   1637  & 671\\ 
                      & Male &      1655 & 722\\ 
 Other                & Female &    113  & 42 \\ 
                      & Male &       181 & 70\\ 
 White                & Female &   9065  & 3962 \\ 
                      & Male &     20117 & 8618 \\ 
\bottomrule
\end{tabular}
\caption{Adult group sizes.}\label{tabPopAdult}
\end{table}

\begin{table}[h!]
\centering
\small
\begin{tabular}{c c c c}

Race   & Sex & Train & Test \\ 
\toprule
Alaska Native      & Female &     2 & 0\\ 
American Indian    & Female &    97 & 47\\ 
                   & Male &       75 & 56\\ 
Amerindian Tribes  & Female &    25 & 12\\ 
                   & Male &       31 & 10\\ 
Asian              & Female &  2178 & 965\\ 
                   & Male &     1887 & 847\\ 
Black              & Female &   618 & 283\\ 
                   & Male &      645 & 290\\ 
Native Hawaiian    & Female &    45 & 14\\ 
                   & Male &       52 & 21\\ 
Some Other Race    & Female &  1493 & 675\\ 
                   & Male &     1629 & 697\\ 
Two or More Races  & Female &   692 & 259\\ 
                   & Male &      725 & 309\\ 
White              & Female &  8277 & 3499\\ 
                   & Male &     8046 & 3381\\
\bottomrule
\end{tabular}
\caption{ACS Income group sizes.}\label{tabPopAcsInc}
\end{table}

\begin{table}[h!]
\centering
\small
\begin{tabular}{c c c c}
Race   & Sex & Train & Test \\ 
\toprule

Alaska Native      & Female &     3  & 1\\ 
American Indian    & Female &   109  & 53\\ 
                   & Male &      82  & 35\\ 
Amerindian Tribes  & Female &    37  & 15\\ 
                   & Male &      33  & 16\\ 
Asian              & Female &  2199  & 909\\ 
                   & Male &     1956 & 844\\ 
Black              & Female &   644  & 312 \\ 
                   & Male &      625 & 283\\ 
Native Hawaiian    & Female &    33  & 16\\ 
                   & Male &       39 & 20\\ 
Some Other Race    & Female &  1520  & 683\\ 
                   & Male &     1538 & 672\\ 
Two or More Races  & Female &   688  & 286\\ 
                   & Male &      700 & 267\\ 
White              & Female &  8345  & 3520\\ 
                   & Male &     7966 & 3433\\ 
\bottomrule
\end{tabular}
\caption{ACS Employment group sizes.}\label{tabPopEmpl}
\end{table}

\begin{table}[h!]
\centering
\small
\begin{tabular}{c c c c}

Race   & Sex & Train & Test \\ 
\toprule

Black  & Female &   753   & 314\\ 
       & Male &      518  & 205\\ 
    \addlinespace  \hline  \addlinespace       
Other  & Female &   887   & 415\\ 
       & Male &     1072  & 451\\ 
    \addlinespace  \hline  \addlinespace       
White  & Female &  6493   & 2816\\ 
       & Male &     8862  & 3765\\ 
\bottomrule
\end{tabular}
\caption{LSAC group sizes.}\label{tabPopLSAC}
\end{table}

\begin{table}[h!]
\centering
\small
\begin{tabular}{c c c c}

Race   & Sex & Train & Test \\ 
\toprule
Black  & Female &   453  & 199\\ 
       & Male &     2113 & 931 \\ 
    \addlinespace  \hline  \addlinespace    
Other  & Female &   131  & 45\\ 
       & Male &      639 & 249\\ 
    \addlinespace  \hline  \addlinespace
White  & Female &   391  & 176\\ 
       & Male &     1322 & 564\\ 
\bottomrule
\end{tabular}
\caption{Compas group sizes.}\label{tabPopCompas}
\end{table}

\clearpage
\section{Intersection Metrics}\label{app:intersection}
In this section, we present several group metrics for all datasets. The decision threshold has been $0.5$ for all threshold dependent metrics.

\begin{table*}
\centering
\small
\begin{tabular}{cc ccc}
    \toprule
    Metric & Group &  Baseline & Baseline + DP & Best DP Model \\
    \midrule
Accuracy & American Indian-Female  & 0.7615 $\pm$ 0.0698 & 0.6146 $\pm$ 0.0688 & 0.6830 $\pm$ 0.0149 \\
Accuracy & American Indian-Male  & 0.8154 $\pm$ 0.0671 & 0.7686 $\pm$ 0.0755 & 0.7571 $\pm$ 0.0179 \\
Accuracy & Amerindian Tribes-Female  & 0.7077 $\pm$ 0.0487 & 0.7143 $\pm$ 0.1650 & 0.7636 $\pm$ 0.0469 \\
Accuracy & Amerindian Tribes-Male  & 0.7643 $\pm$ 0.0894 & 0.7000 $\pm$ 0.0704 & 0.7692 $\pm$ 0.0000 \\
Accuracy & Asian-Female  & 0.7615 $\pm$ 0.0090 & 0.7063 $\pm$ 0.0091 & 0.7473 $\pm$ 0.0052 \\
Accuracy & Asian-Male  & 0.8251 $\pm$ 0.0082 & 0.7704 $\pm$ 0.0080 & 0.8355 $\pm$ 0.0047 \\
Accuracy & Black-Female  & 0.7502 $\pm$ 0.0140 & 0.7287 $\pm$ 0.0230 & 0.8171 $\pm$ 0.0055 \\
Accuracy & Black-Male  & 0.8365 $\pm$ 0.0138 & 0.7349 $\pm$ 0.0250 & 0.8354 $\pm$ 0.0077 \\
Accuracy & Native Hawaiian-Female  & 0.6462 $\pm$ 0.1214 & 0.6933 $\pm$ 0.1142 & 0.8636 $\pm$ 0.0000 \\
Accuracy & Native Hawaiian-Male  & 0.8375 $\pm$ 0.0437 & 0.8300 $\pm$ 0.0888 & 0.8000 $\pm$ 0.0471 \\
Accuracy & Some Other Race-Female  & 0.7521 $\pm$ 0.0088 & 0.7285 $\pm$ 0.0050 & 0.7725 $\pm$ 0.0058 \\
Accuracy & Some Other Race-Male  & 0.8477 $\pm$ 0.0079 & 0.8084 $\pm$ 0.0074 & 0.8612 $\pm$ 0.0081 \\
Accuracy & Two or More Races-Female  & 0.8026 $\pm$ 0.0181 & 0.7907 $\pm$ 0.0100 & 0.8139 $\pm$ 0.0057 \\
Accuracy & Two or More Races-Male  & 0.8911 $\pm$ 0.0124 & 0.8132 $\pm$ 0.0106 & 0.8920 $\pm$ 0.0037 \\
Accuracy & White-Female  & 0.7743 $\pm$ 0.0026 & 0.7025 $\pm$ 0.0064 & 0.7624 $\pm$ 0.0018 \\
Accuracy & White-Male  & 0.8283 $\pm$ 0.0052 & 0.7724 $\pm$ 0.0066 & 0.8283 $\pm$ 0.0018 \\
    \addlinespace  \hline  \addlinespace
Precision & American Indian-Female  & 0.6486 $\pm$ 0.0889 & 0.5160 $\pm$ 0.0839 & 0.5924 $\pm$ 0.0158 \\
Precision & American Indian-Male  & 0.8531 $\pm$ 0.1051 & 0.8285 $\pm$ 0.1193 & 0.6816 $\pm$ 0.0213 \\
Precision & Amerindian Tribes-Female  & 0.4150 $\pm$ 0.0747 & 0.6933 $\pm$ 0.2779 & 0.8893 $\pm$ 0.0588 \\
Precision & Amerindian Tribes-Male  & 0.9650 $\pm$ 0.0568 & 0.4886 $\pm$ 0.1907 & 0.7143 $\pm$ 0.0000 \\
Precision & Asian-Female  & 0.7185 $\pm$ 0.0136 & 0.6305 $\pm$ 0.0084 & 0.6626 $\pm$ 0.0055 \\
Precision & Asian-Male  & 0.7978 $\pm$ 0.0087 & 0.7103 $\pm$ 0.0093 & 0.8048 $\pm$ 0.0047 \\
Precision & Black-Female  & 0.6805 $\pm$ 0.0208 & 0.5546 $\pm$ 0.0256 & 0.7496 $\pm$ 0.0090 \\
Precision & Black-Male  & 0.7437 $\pm$ 0.0232 & 0.5884 $\pm$ 0.0247 & 0.7313 $\pm$ 0.0072 \\
Precision & Native Hawaiian-Female  & 0.6855 $\pm$ 0.0944 & 0.6640 $\pm$ 0.1147 & 0.8333 $\pm$ 0.0000 \\
Precision & Native Hawaiian-Male  & 0.8086 $\pm$ 0.0479 & 0.8271 $\pm$ 0.0969 & 0.7848 $\pm$ 0.0155 \\
Precision & Some Other Race-Female  & 0.6797 $\pm$ 0.0123 & 0.6241 $\pm$ 0.0046 & 0.6649 $\pm$ 0.0057 \\
Precision & Some Other Race-Male  & 0.8146 $\pm$ 0.0073 & 0.7457 $\pm$ 0.0103 & 0.8217 $\pm$ 0.0057 \\
Precision & Two or More Races-Female  & 0.7170 $\pm$ 0.0233 & 0.6398 $\pm$ 0.0152 & 0.6713 $\pm$ 0.0092 \\
Precision & Two or More Races-Male  & 0.8303 $\pm$ 0.0166 & 0.6430 $\pm$ 0.0194 & 0.7942 $\pm$ 0.0065 \\
Precision & White-Female  & 0.7132 $\pm$ 0.0044 & 0.6032 $\pm$ 0.0069 & 0.6755 $\pm$ 0.0026 \\
Precision & White-Male  & 0.8045 $\pm$ 0.0065 & 0.7100 $\pm$ 0.0072 & 0.8087 $\pm$ 0.0062 \\
    \addlinespace  \hline  \addlinespace
TNR & American Indian-Female  & 0.8059 $\pm$ 0.0623 & 0.5800 $\pm$ 0.1860 & 0.6290 $\pm$ 0.0228 \\
TNR & American Indian-Male  & 0.9292 $\pm$ 0.0623 & 0.8222 $\pm$ 0.1589 & 0.6704 $\pm$ 0.0324 \\
TNR & Amerindian Tribes-Female  & 0.7100 $\pm$ 0.0316 & 0.6333 $\pm$ 0.2919 & 0.7333 $\pm$ 0.1405 \\
TNR & Amerindian Tribes-Male  & 0.9000 $\pm$ 0.1610 & 0.6692 $\pm$ 0.2085 & 0.7143 $\pm$ 0.0000 \\
TNR & Asian-Female  & 0.7206 $\pm$ 0.0230 & 0.5112 $\pm$ 0.0224 & 0.6251 $\pm$ 0.0126 \\
TNR & Asian-Male  & 0.7654 $\pm$ 0.0139 & 0.5591 $\pm$ 0.0228 & 0.7357 $\pm$ 0.0076 \\
TNR & Black-Female  & 0.7538 $\pm$ 0.0265 & 0.6707 $\pm$ 0.0460 & 0.8053 $\pm$ 0.0106 \\
TNR & Black-Male  & 0.8435 $\pm$ 0.0180 & 0.6506 $\pm$ 0.0439 & 0.8093 $\pm$ 0.0060 \\
TNR & Native Hawaiian-Female  & 0.4200 $\pm$ 0.1751 & 0.6625 $\pm$ 0.2503 & 0.8182 $\pm$ 0.0000 \\
TNR & Native Hawaiian-Male  & 0.7875 $\pm$ 0.0604 & 0.8364 $\pm$ 0.1271 & 0.6250 $\pm$ 0.0000 \\
TNR & Some Other Race-Female  & 0.7355 $\pm$ 0.0180 & 0.6016 $\pm$ 0.0194 & 0.6648 $\pm$ 0.0085 \\
TNR & Some Other Race-Male  & 0.7864 $\pm$ 0.0096 & 0.6851 $\pm$ 0.0213 & 0.8069 $\pm$ 0.0079 \\
TNR & Two or More Races-Female  & 0.8150 $\pm$ 0.0192 & 0.7764 $\pm$ 0.0302 & 0.7790 $\pm$ 0.0102 \\
TNR & Two or More Races-Male  & 0.8620 $\pm$ 0.0162 & 0.7832 $\pm$ 0.0256 & 0.8784 $\pm$ 0.0058 \\
TNR & White-Female  & 0.7551 $\pm$ 0.0096 & 0.5605 $\pm$ 0.0182 & 0.6946 $\pm$ 0.0070 \\
TNR & White-Male  & 0.8011 $\pm$ 0.0090 & 0.6374 $\pm$ 0.0141 & 0.7930 $\pm$ 0.0107 \\
    \bottomrule
\end{tabular}
\caption{ACS Employment: Mean and standard deviation values for 10 training runs for the best model configurations with and without differential privacy for several metrics regarding all subgroups between race and gender. More metrics for this dataset in Table~\ref{tabIntersectionMetricsACSEmploy2}.}\label{tabIntersectionMetricsACSEmploy1}. 
\end{table*}

\begin{table*}
\centering
\small
\begin{tabular}{cc ccc}
    \toprule
    Metric & Group &  Baseline & Baseline + DP & Best DP Model \\
    \midrule
TPR & American Indian-Female  & 0.6778 $\pm$ 0.1610 & 0.6687 $\pm$ 0.2166 & 0.7591 $\pm$ 0.0220 \\
TPR & American Indian-Male  & 0.6333 $\pm$ 0.1305 & 0.7118 $\pm$ 0.2191 & 0.8636 $\pm$ 0.0000 \\
TPR & Amerindian Tribes-Female  & 0.7000 $\pm$ 0.1892 & 0.7750 $\pm$ 0.3810 & 0.7750 $\pm$ 0.0527 \\
TPR & Amerindian Tribes-Male  & 0.7273 $\pm$ 0.0958 & 0.7667 $\pm$ 0.3258 & 0.8333 $\pm$ 0.0000 \\
TPR & Asian-Female  & 0.8080 $\pm$ 0.0227 & 0.9221 $\pm$ 0.0189 & 0.8961 $\pm$ 0.0150 \\
TPR & Asian-Male  & 0.8821 $\pm$ 0.0165 & 0.9578 $\pm$ 0.0107 & 0.9198 $\pm$ 0.0050 \\
TPR & Black-Female  & 0.7450 $\pm$ 0.0322 & 0.8500 $\pm$ 0.0400 & 0.8339 $\pm$ 0.0134 \\
TPR & Black-Male  & 0.8236 $\pm$ 0.0150 & 0.8850 $\pm$ 0.0467 & 0.8796 $\pm$ 0.0190 \\
TPR & Native Hawaiian-Female  & 0.7875 $\pm$ 0.1324 & 0.7286 $\pm$ 0.2970 & 0.9091 $\pm$ 0.0000 \\
TPR & Native Hawaiian-Male  & 0.8875 $\pm$ 0.0710 & 0.8222 $\pm$ 0.2468 & 0.9167 $\pm$ 0.0786 \\
TPR & Some Other Race-Female  & 0.7751 $\pm$ 0.0216 & 0.9015 $\pm$ 0.0318 & 0.9218 $\pm$ 0.0102 \\
TPR & Some Other Race-Male  & 0.9070 $\pm$ 0.0114 & 0.9331 $\pm$ 0.0192 & 0.9171 $\pm$ 0.0172 \\
TPR & Two or More Races-Female  & 0.7820 $\pm$ 0.0343 & 0.8202 $\pm$ 0.0736 & 0.8822 $\pm$ 0.0078 \\
TPR & Two or More Races-Male  & 0.9314 $\pm$ 0.0209 & 0.8810 $\pm$ 0.0478 & 0.9186 $\pm$ 0.0133 \\
TPR & White-Female  & 0.7994 $\pm$ 0.0157 & 0.8923 $\pm$ 0.0175 & 0.8535 $\pm$ 0.0114 \\
TPR & White-Male  & 0.8569 $\pm$ 0.0104 & 0.9110 $\pm$ 0.0116 & 0.8632 $\pm$ 0.0112 \\
    \addlinespace  \hline  \addlinespace
Selection Rate & American Indian-Female  & 0.3615 $\pm$ 0.0683 & 0.5171 $\pm$ 0.1878 & 0.5321 $\pm$ 0.0173 \\
Selection Rate & American Indian-Male  & 0.2872 $\pm$ 0.0590 & 0.4371 $\pm$ 0.1741 & 0.5694 $\pm$ 0.0179 \\
Selection Rate & Amerindian Tribes-Female  & 0.3846 $\pm$ 0.0513 & 0.6000 $\pm$ 0.3144 & 0.6364 $\pm$ 0.0606 \\
Selection Rate & Amerindian Tribes-Male  & 0.5929 $\pm$ 0.0757 & 0.4684 $\pm$ 0.2386 & 0.5385 $\pm$ 0.0000 \\
Selection Rate & Asian-Female  & 0.5270 $\pm$ 0.0211 & 0.6945 $\pm$ 0.0188 & 0.6099 $\pm$ 0.0126 \\
Selection Rate & Asian-Male  & 0.5661 $\pm$ 0.0129 & 0.7149 $\pm$ 0.0151 & 0.6198 $\pm$ 0.0041 \\
Selection Rate & Black-Female  & 0.4519 $\pm$ 0.0253 & 0.4980 $\pm$ 0.0417 & 0.4575 $\pm$ 0.0104 \\
Selection Rate & Black-Male  & 0.3932 $\pm$ 0.0116 & 0.5421 $\pm$ 0.0390 & 0.4464 $\pm$ 0.0083 \\
Selection Rate & Native Hawaiian-Female  & 0.7077 $\pm$ 0.0873 & 0.5200 $\pm$ 0.2471 & 0.5455 $\pm$ 0.0000 \\
Selection Rate & Native Hawaiian-Male  & 0.5500 $\pm$ 0.0493 & 0.4600 $\pm$ 0.1630 & 0.7000 $\pm$ 0.0471 \\
Selection Rate & Some Other Race-Female  & 0.4788 $\pm$ 0.0175 & 0.6112 $\pm$ 0.0243 & 0.5810 $\pm$ 0.0072 \\
Selection Rate & Some Other Race-Male  & 0.5662 $\pm$ 0.0070 & 0.6222 $\pm$ 0.0189 & 0.5495 $\pm$ 0.0105 \\
Selection Rate & Two or More Races-Female  & 0.4086 $\pm$ 0.0171 & 0.4176 $\pm$ 0.0433 & 0.4447 $\pm$ 0.0085 \\
Selection Rate & Two or More Races-Male  & 0.4712 $\pm$ 0.0133 & 0.4202 $\pm$ 0.0308 & 0.3909 $\pm$ 0.0075 \\
Selection Rate & White-Female  & 0.4846 $\pm$ 0.0120 & 0.6334 $\pm$ 0.0169 & 0.5394 $\pm$ 0.0087 \\
Selection Rate & White-Male  & 0.5203 $\pm$ 0.0082 & 0.6332 $\pm$ 0.0112 & 0.5372 $\pm$ 0.0108 \\
    \addlinespace  \hline  \addlinespace
ROC AUC & American Indian-Female  & 0.7418 $\pm$ 0.0870 & 0.6244 $\pm$ 0.0658 & 0.6941 $\pm$ 0.0145 \\
ROC AUC & American Indian-Male  & 0.7812 $\pm$ 0.0760 & 0.7670 $\pm$ 0.0775 & 0.7670 $\pm$ 0.0162 \\
ROC AUC & Amerindian Tribes-Female  & 0.7050 $\pm$ 0.0949 & 0.7042 $\pm$ 0.1449 & 0.7542 $\pm$ 0.0686 \\
ROC AUC & Amerindian Tribes-Male  & 0.8136 $\pm$ 0.1023 & 0.7179 $\pm$ 0.0857 & 0.7738 $\pm$ 0.0000 \\
ROC AUC & Asian-Female  & 0.7643 $\pm$ 0.0088 & 0.7167 $\pm$ 0.0088 & 0.7606 $\pm$ 0.0054 \\
ROC AUC & Asian-Male  & 0.8237 $\pm$ 0.0082 & 0.7585 $\pm$ 0.0087 & 0.8277 $\pm$ 0.0048 \\
ROC AUC & Black-Female  & 0.7494 $\pm$ 0.0143 & 0.7603 $\pm$ 0.0154 & 0.8196 $\pm$ 0.0057 \\
ROC AUC & Black-Male  & 0.8336 $\pm$ 0.0129 & 0.7678 $\pm$ 0.0233 & 0.8445 $\pm$ 0.0097 \\
ROC AUC & Native Hawaiian-Female  & 0.6038 $\pm$ 0.1257 & 0.6955 $\pm$ 0.1167 & 0.8636 $\pm$ 0.0000 \\
ROC AUC & Native Hawaiian-Male  & 0.8375 $\pm$ 0.0437 & 0.8293 $\pm$ 0.0992 & 0.7708 $\pm$ 0.0393 \\
ROC AUC & Some Other Race-Female  & 0.7553 $\pm$ 0.0090 & 0.7515 $\pm$ 0.0077 & 0.7933 $\pm$ 0.0059 \\
ROC AUC & Some Other Race-Male  & 0.8467 $\pm$ 0.0079 & 0.8091 $\pm$ 0.0073 & 0.8620 $\pm$ 0.0082 \\
ROC AUC & Two or More Races-Female  & 0.7985 $\pm$ 0.0201 & 0.7983 $\pm$ 0.0239 & 0.8306 $\pm$ 0.0043 \\
ROC AUC & Two or More Races-Male  & 0.8967 $\pm$ 0.0127 & 0.8321 $\pm$ 0.0156 & 0.8985 $\pm$ 0.0053 \\
ROC AUC & White-Female  & 0.7773 $\pm$ 0.0038 & 0.7264 $\pm$ 0.0058 & 0.7740 $\pm$ 0.0027 \\
ROC AUC & White-Male  & 0.8290 $\pm$ 0.0052 & 0.7742 $\pm$ 0.0065 & 0.8281 $\pm$ 0.0018 \\
    \bottomrule
\end{tabular}
\caption{ACS Employment: Mean and standard deviation values for 10 training runs for the best model configurations with and without differential privacy for several metrics regarding all subgroups between race and gender.}\label{tabIntersectionMetricsACSEmploy2}
\end{table*}

\begin{table*}
\centering
\small
\begin{tabular}{cc ccc}
    \toprule
    Metric & Group &  Baseline & Baseline + DP & Best DP Model \\
    \midrule
Accuracy & American Indian-Female  & 0.7121 $\pm$ 0.0434 & 0.7420 $\pm$ 0.0394 & 0.7814 $\pm$ 0.0196 \\
Accuracy & American Indian-Male  & 0.8636 $\pm$ 0.0262 & 0.7323 $\pm$ 0.0505 & 0.7605 $\pm$ 0.0083 \\
Accuracy & Amerindian Tribes-Female  & 0.8000 $\pm$ 0.0493 & 0.6000 $\pm$ 0.1042 & 0.7308 $\pm$ 0.0405 \\
Accuracy & Amerindian Tribes-Male  & 0.7167 $\pm$ 0.0315 & 0.8250 $\pm$ 0.1344 & 0.8700 $\pm$ 0.0483 \\
Accuracy & Asian-Female  & 0.7709 $\pm$ 0.0056 & 0.6838 $\pm$ 0.0063 & 0.7775 $\pm$ 0.0054 \\
Accuracy & Asian-Male  & 0.8307 $\pm$ 0.0053 & 0.7741 $\pm$ 0.0051 & 0.8580 $\pm$ 0.0039 \\
Accuracy & Black-Female  & 0.8067 $\pm$ 0.0133 & 0.7209 $\pm$ 0.0056 & 0.7925 $\pm$ 0.0059 \\
Accuracy & Black-Male  & 0.8221 $\pm$ 0.0128 & 0.7701 $\pm$ 0.0282 & 0.8183 $\pm$ 0.0056 \\
Accuracy & Native Hawaiian-Female  & 0.7381 $\pm$ 0.0645 & 0.7733 $\pm$ 0.1891 & 0.8588 $\pm$ 0.0304 \\
Accuracy & Native Hawaiian-Male  & 0.8765 $\pm$ 0.0585 & 0.7652 $\pm$ 0.0622 & 0.9000 $\pm$ 0.0323 \\
Accuracy & Some Other Race-Female  & 0.7401 $\pm$ 0.0094 & 0.7137 $\pm$ 0.0056 & 0.7658 $\pm$ 0.0023 \\
Accuracy & Some Other Race-Male  & 0.8592 $\pm$ 0.0054 & 0.8111 $\pm$ 0.0080 & 0.8554 $\pm$ 0.0046 \\
Accuracy & Two or More Races-Female  & 0.8204 $\pm$ 0.0153 & 0.7599 $\pm$ 0.0158 & 0.8063 $\pm$ 0.0044 \\
Accuracy & Two or More Races-Male  & 0.8765 $\pm$ 0.0082 & 0.8537 $\pm$ 0.0135 & 0.8663 $\pm$ 0.0026 \\
Accuracy & White-Female  & 0.7704 $\pm$ 0.0025 & 0.7038 $\pm$ 0.0098 & 0.7759 $\pm$ 0.0023 \\
Accuracy & White-Male  & 0.8337 $\pm$ 0.0038 & 0.7783 $\pm$ 0.0046 & 0.8447 $\pm$ 0.0014 \\
    \addlinespace  \hline  \addlinespace
Precision & American Indian-Female  & 0.6203 $\pm$ 0.0361 & 0.7015 $\pm$ 0.0713 & 0.6241 $\pm$ 0.0248 \\
Precision & American Indian-Male  & 0.8072 $\pm$ 0.0595 & 0.6598 $\pm$ 0.0831 & 0.6789 $\pm$ 0.0094 \\
Precision & Amerindian Tribes-Female  & 0.7561 $\pm$ 0.0372 & 0.3751 $\pm$ 0.2904 & 0.5982 $\pm$ 0.0282 \\
Precision & Amerindian Tribes-Male  & 0.6183 $\pm$ 0.0247 & 0.5067 $\pm$ 0.4133 & 0.7250 $\pm$ 0.0403 \\
Precision & Asian-Female  & 0.7212 $\pm$ 0.0090 & 0.6068 $\pm$ 0.0061 & 0.6896 $\pm$ 0.0054 \\
Precision & Asian-Male  & 0.8005 $\pm$ 0.0055 & 0.7051 $\pm$ 0.0073 & 0.8287 $\pm$ 0.0044 \\
Precision & Black-Female  & 0.7261 $\pm$ 0.0157 & 0.6286 $\pm$ 0.0115 & 0.7264 $\pm$ 0.0094 \\
Precision & Black-Male  & 0.6692 $\pm$ 0.0143 & 0.6266 $\pm$ 0.0374 & 0.7198 $\pm$ 0.0064 \\
Precision & Native Hawaiian-Female  & 0.7276 $\pm$ 0.0460 & 0.8000 $\pm$ 0.4216 & 0.6827 $\pm$ 0.0425 \\
Precision & Native Hawaiian-Male  & 0.9549 $\pm$ 0.0390 & 0.5415 $\pm$ 0.1072 & 0.7929 $\pm$ 0.0553 \\
Precision & Some Other Race-Female  & 0.6577 $\pm$ 0.0143 & 0.6113 $\pm$ 0.0087 & 0.6403 $\pm$ 0.0016 \\
Precision & Some Other Race-Male  & 0.8289 $\pm$ 0.0078 & 0.7432 $\pm$ 0.0097 & 0.8093 $\pm$ 0.0025 \\
Precision & Two or More Races-Female  & 0.7178 $\pm$ 0.0200 & 0.6210 $\pm$ 0.0173 & 0.6932 $\pm$ 0.0053 \\
Precision & Two or More Races-Male  & 0.7755 $\pm$ 0.0138 & 0.7547 $\pm$ 0.0223 & 0.7725 $\pm$ 0.0032 \\
Precision & White-Female  & 0.7000 $\pm$ 0.0062 & 0.6037 $\pm$ 0.0099 & 0.6901 $\pm$ 0.0024 \\
Precision & White-Male  & 0.8075 $\pm$ 0.0049 & 0.7165 $\pm$ 0.0054 & 0.8268 $\pm$ 0.0028 \\
    \bottomrule
\end{tabular}
\caption{ACS Income: Mean and standard deviation values for 10 training runs for the best model configurations with and without differential privacy for Accuracy and Precision on all subgroups between race and gender.}
\end{table*}

\begin{table*}
\centering
\small
\begin{tabular}{cc ccc}
    \toprule
    Metric & Group &  Baseline & Baseline + DP & Best DP Model \\
    \midrule
TNR & American Indian-Female  & 0.7300 $\pm$ 0.0422 & 0.8033 $\pm$ 0.0838 & 0.7586 $\pm$ 0.0230 \\
TNR & American Indian-Male  & 0.8821 $\pm$ 0.0478 & 0.6778 $\pm$ 0.1304 & 0.6619 $\pm$ 0.0151 \\
TNR & Amerindian Tribes-Female  & 0.7125 $\pm$ 0.0604 & 0.6600 $\pm$ 0.2633 & 0.6250 $\pm$ 0.0000 \\
TNR & Amerindian Tribes-Male  & 0.5300 $\pm$ 0.0483 & 0.8833 $\pm$ 0.1766 & 0.8571 $\pm$ 0.0000 \\
TNR & Asian-Female  & 0.7400 $\pm$ 0.0167 & 0.4738 $\pm$ 0.0195 & 0.6454 $\pm$ 0.0086 \\
TNR & Asian-Male  & 0.7431 $\pm$ 0.0101 & 0.5744 $\pm$ 0.0184 & 0.7615 $\pm$ 0.0085 \\
TNR & Black-Female  & 0.8139 $\pm$ 0.0128 & 0.6442 $\pm$ 0.0322 & 0.7580 $\pm$ 0.0130 \\
TNR & Black-Male  & 0.8114 $\pm$ 0.0112 & 0.7195 $\pm$ 0.0460 & 0.7913 $\pm$ 0.0064 \\
TNR & Native Hawaiian-Female  & 0.5667 $\pm$ 0.0820 & 1.0000 $\pm$ 0.0000 & 0.8083 $\pm$ 0.0403 \\
TNR & Native Hawaiian-Male  & 0.8000 $\pm$ 0.1721 & 0.7471 $\pm$ 0.1002 & 0.8400 $\pm$ 0.0516 \\
TNR & Some Other Race-Female  & 0.7508 $\pm$ 0.0200 & 0.5974 $\pm$ 0.0291 & 0.6631 $\pm$ 0.0027 \\
TNR & Some Other Race-Male  & 0.8040 $\pm$ 0.0116 & 0.7020 $\pm$ 0.0181 & 0.7704 $\pm$ 0.0033 \\
TNR & Two or More Races-Female  & 0.8206 $\pm$ 0.0168 & 0.7529 $\pm$ 0.0230 & 0.7677 $\pm$ 0.0053 \\
TNR & Two or More Races-Male  & 0.8627 $\pm$ 0.0103 & 0.8574 $\pm$ 0.0200 & 0.8481 $\pm$ 0.0029 \\
TNR & White-Female  & 0.7547 $\pm$ 0.0104 & 0.5747 $\pm$ 0.0202 & 0.7074 $\pm$ 0.0041 \\
TNR & White-Male  & 0.7902 $\pm$ 0.0075 & 0.6462 $\pm$ 0.0125 & 0.8172 $\pm$ 0.0042 \\
    \addlinespace  \hline  \addlinespace
TPR & American Indian-Female  & 0.6846 $\pm$ 0.1330 & 0.6500 $\pm$ 0.1810 & 0.8286 $\pm$ 0.0369 \\
TPR & American Indian-Male  & 0.8313 $\pm$ 0.0836 & 0.8077 $\pm$ 0.1462 & 0.8824 $\pm$ 0.0000 \\
TPR & Amerindian Tribes-Female  & 0.8875 $\pm$ 0.0922 & 0.4800 $\pm$ 0.4022 & 0.9000 $\pm$ 0.1054 \\
TPR & Amerindian Tribes-Male  & 0.9500 $\pm$ 0.0645 & 0.6500 $\pm$ 0.4743 & 0.9000 $\pm$ 0.1610 \\
TPR & Asian-Female  & 0.8080 $\pm$ 0.0218 & 0.9224 $\pm$ 0.0155 & 0.9341 $\pm$ 0.0055 \\
TPR & Asian-Male  & 0.9078 $\pm$ 0.0116 & 0.9632 $\pm$ 0.0087 & 0.9364 $\pm$ 0.0089 \\
TPR & Black-Female  & 0.7951 $\pm$ 0.0261 & 0.8264 $\pm$ 0.0350 & 0.8374 $\pm$ 0.0101 \\
TPR & Black-Male  & 0.8458 $\pm$ 0.0389 & 0.8639 $\pm$ 0.0393 & 0.8617 $\pm$ 0.0123 \\
TPR & Native Hawaiian-Female  & 0.8667 $\pm$ 0.0805 & 0.5750 $\pm$ 0.3545 & 0.9800 $\pm$ 0.0632 \\
TPR & Native Hawaiian-Male  & 0.8929 $\pm$ 0.0505 & 0.8167 $\pm$ 0.2659 & 1.0000 $\pm$ 0.0000 \\
TPR & Some Other Race-Female  & 0.7237 $\pm$ 0.0283 & 0.8746 $\pm$ 0.0364 & 0.9241 $\pm$ 0.0076 \\
TPR & Some Other Race-Male  & 0.9122 $\pm$ 0.0099 & 0.9287 $\pm$ 0.0168 & 0.9371 $\pm$ 0.0089 \\
TPR & Two or More Races-Female  & 0.8200 $\pm$ 0.0346 & 0.7733 $\pm$ 0.0581 & 0.8702 $\pm$ 0.0069 \\
TPR & Two or More Races-Male  & 0.9028 $\pm$ 0.0090 & 0.8464 $\pm$ 0.0417 & 0.8981 $\pm$ 0.0067 \\
TPR & White-Female  & 0.7921 $\pm$ 0.0116 & 0.8791 $\pm$ 0.0127 & 0.8672 $\pm$ 0.0066 \\
TPR & White-Male  & 0.8771 $\pm$ 0.0084 & 0.9132 $\pm$ 0.0127 & 0.8722 $\pm$ 0.0041 \\
    \bottomrule
\end{tabular}
\caption{ACS Income: Mean and standard deviation values for 10 training runs for the best model configurations with and without differential privacy for the True Negative Rate and True Positive Rate on all subgroups between race and gender.}
\end{table*}

\begin{table*}
\centering
\small
\begin{tabular}{cc ccc}
    \toprule
    Metric & Group &  Baseline & Baseline + DP & Best DP Model \\
    \midrule
Selection Rate & American Indian-Female  & 0.4333 $\pm$ 0.0701 & 0.3780 $\pm$ 0.1183 & 0.4326 $\pm$ 0.0196 \\
Selection Rate & American Indian-Male  & 0.3773 $\pm$ 0.0548 & 0.5258 $\pm$ 0.1282 & 0.5816 $\pm$ 0.0083 \\
Selection Rate & Amerindian Tribes-Female  & 0.5875 $\pm$ 0.0604 & 0.3867 $\pm$ 0.2945 & 0.5769 $\pm$ 0.0405 \\
Selection Rate & Amerindian Tribes-Male  & 0.6833 $\pm$ 0.0457 & 0.2500 $\pm$ 0.2125 & 0.3700 $\pm$ 0.0483 \\
Selection Rate & Asian-Female  & 0.5087 $\pm$ 0.0182 & 0.7116 $\pm$ 0.0167 & 0.6197 $\pm$ 0.0053 \\
Selection Rate & Asian-Male  & 0.6029 $\pm$ 0.0096 & 0.7017 $\pm$ 0.0132 & 0.6237 $\pm$ 0.0079 \\
Selection Rate & Black-Female  & 0.4193 $\pm$ 0.0121 & 0.5539 $\pm$ 0.0331 & 0.5004 $\pm$ 0.0106 \\
Selection Rate & Black-Male  & 0.3929 $\pm$ 0.0157 & 0.4851 $\pm$ 0.0370 & 0.4591 $\pm$ 0.0066 \\
Selection Rate & Native Hawaiian-Female  & 0.6810 $\pm$ 0.0504 & 0.3067 $\pm$ 0.1891 & 0.4235 $\pm$ 0.0372 \\
Selection Rate & Native Hawaiian-Male  & 0.7706 $\pm$ 0.0434 & 0.4000 $\pm$ 0.1293 & 0.4750 $\pm$ 0.0323 \\
Selection Rate & Some Other Race-Female  & 0.4379 $\pm$ 0.0213 & 0.6006 $\pm$ 0.0317 & 0.5680 $\pm$ 0.0042 \\
Selection Rate & Some Other Race-Male  & 0.5612 $\pm$ 0.0093 & 0.6015 $\pm$ 0.0156 & 0.5903 $\pm$ 0.0050 \\
Selection Rate & Two or More Races-Female  & 0.4082 $\pm$ 0.0175 & 0.4279 $\pm$ 0.0317 & 0.4723 $\pm$ 0.0040 \\
Selection Rate & Two or More Races-Male  & 0.4006 $\pm$ 0.0066 & 0.3821 $\pm$ 0.0238 & 0.4242 $\pm$ 0.0035 \\
Selection Rate & White-Female  & 0.4746 $\pm$ 0.0107 & 0.6178 $\pm$ 0.0152 & 0.5391 $\pm$ 0.0046 \\
Selection Rate & White-Male  & 0.5441 $\pm$ 0.0070 & 0.6305 $\pm$ 0.0117 & 0.5276 $\pm$ 0.0039 \\
    \addlinespace  \hline  \addlinespace
ROC AUC & American Indian-Female  & 0.7073 $\pm$ 0.0573 & 0.7267 $\pm$ 0.0589 & 0.7936 $\pm$ 0.0217 \\
ROC AUC & American Indian-Male  & 0.8567 $\pm$ 0.0326 & 0.7427 $\pm$ 0.0497 & 0.7721 $\pm$ 0.0075 \\
ROC AUC & Amerindian Tribes-Female  & 0.8000 $\pm$ 0.0493 & 0.5700 $\pm$ 0.1229 & 0.7625 $\pm$ 0.0527 \\
ROC AUC & Amerindian Tribes-Male  & 0.7400 $\pm$ 0.0327 & 0.7667 $\pm$ 0.2144 & 0.8786 $\pm$ 0.0805 \\
ROC AUC & Asian-Female  & 0.7740 $\pm$ 0.0061 & 0.6981 $\pm$ 0.0058 & 0.7898 $\pm$ 0.0052 \\
ROC AUC & Asian-Male  & 0.8254 $\pm$ 0.0052 & 0.7688 $\pm$ 0.0055 & 0.8489 $\pm$ 0.0038 \\
ROC AUC & Black-Female  & 0.8045 $\pm$ 0.0151 & 0.7353 $\pm$ 0.0043 & 0.7977 $\pm$ 0.0053 \\
ROC AUC & Black-Male  & 0.8286 $\pm$ 0.0190 & 0.7917 $\pm$ 0.0245 & 0.8265 $\pm$ 0.0064 \\
ROC AUC & Native Hawaiian-Female  & 0.7167 $\pm$ 0.0642 & 0.7875 $\pm$ 0.1773 & 0.8942 $\pm$ 0.0336 \\
ROC AUC & Native Hawaiian-Male  & 0.8464 $\pm$ 0.0968 & 0.7819 $\pm$ 0.1088 & 0.9200 $\pm$ 0.0258 \\
ROC AUC & Some Other Race-Female  & 0.7373 $\pm$ 0.0104 & 0.7360 $\pm$ 0.0065 & 0.7936 $\pm$ 0.0031 \\
ROC AUC & Some Other Race-Male  & 0.8581 $\pm$ 0.0054 & 0.8153 $\pm$ 0.0079 & 0.8537 $\pm$ 0.0046 \\
ROC AUC & Two or More Races-Female  & 0.8203 $\pm$ 0.0182 & 0.7631 $\pm$ 0.0237 & 0.8190 $\pm$ 0.0046 \\
ROC AUC & Two or More Races-Male  & 0.8828 $\pm$ 0.0077 & 0.8519 $\pm$ 0.0179 & 0.8731 $\pm$ 0.0032 \\
ROC AUC & White-Female  & 0.7734 $\pm$ 0.0023 & 0.7269 $\pm$ 0.0086 & 0.7873 $\pm$ 0.0026 \\
ROC AUC & White-Male  & 0.8336 $\pm$ 0.0038 & 0.7797 $\pm$ 0.0046 & 0.8447 $\pm$ 0.0014 \\
    \bottomrule

\end{tabular}
\caption{ACS Income: Mean and standard deviation values for 10 training runs for the best model configurations with and without differential privacy for Selection Rate and ROC AUC on all subgroups between race and gender.}\label{tabIntersectionMetricsACSIncome2}
\end{table*}

\begin{table*}
\centering
\small
\begin{tabular}{cc ccc}
    \toprule
    Metric & Group &  Baseline & Baseline + DP & Best DP Model \\
    \midrule
Accuracy & Black-Female  & 0.7659 $\pm$ 0.0155 & 0.7331 $\pm$ 0.0087 & 0.7379 $\pm$ 0.0078 \\
Accuracy & Black-Male  & 0.7420 $\pm$ 0.0310 & 0.6722 $\pm$ 0.0100 & 0.7005 $\pm$ 0.0181 \\
Accuracy & Other-Female  & 0.8272 $\pm$ 0.0098 & 0.7937 $\pm$ 0.0058 & 0.8099 $\pm$ 0.0110 \\
Accuracy & Other-Male  & 0.8344 $\pm$ 0.0052 & 0.8144 $\pm$ 0.0033 & 0.8233 $\pm$ 0.0072 \\
Accuracy & White-Female  & 0.8656 $\pm$ 0.0030 & 0.8332 $\pm$ 0.0009 & 0.8426 $\pm$ 0.0085 \\
Accuracy & White-Male  & 0.8826 $\pm$ 0.0033 & 0.8521 $\pm$ 0.0026 & 0.8645 $\pm$ 0.0070 \\
    \addlinespace  \hline  \addlinespace
Precision & Black-Female  & 0.7706 $\pm$ 0.0234 & 0.7039 $\pm$ 0.0108 & 0.7108 $\pm$ 0.0067 \\
Precision & Black-Male  & 0.7354 $\pm$ 0.0513 & 0.6311 $\pm$ 0.0063 & 0.6488 $\pm$ 0.0142 \\
Precision & Other-Female  & 0.8241 $\pm$ 0.0119 & 0.7861 $\pm$ 0.0059 & 0.8006 $\pm$ 0.0109 \\
Precision & Other-Male  & 0.8530 $\pm$ 0.0102 & 0.8117 $\pm$ 0.0032 & 0.8216 $\pm$ 0.0087 \\
Precision & White-Female  & 0.8803 $\pm$ 0.0047 & 0.8316 $\pm$ 0.0010 & 0.8486 $\pm$ 0.0142 \\
Precision & White-Male  & 0.9011 $\pm$ 0.0043 & 0.8513 $\pm$ 0.0025 & 0.8685 $\pm$ 0.0109 \\
    \addlinespace  \hline  \addlinespace
TNR & Black-Female  & 0.5960 $\pm$ 0.0753 & 0.3766 $\pm$ 0.0399 & 0.4040 $\pm$ 0.0239 \\
TNR & Black-Male  & 0.6594 $\pm$ 0.0936 & 0.3865 $\pm$ 0.0409 & 0.4135 $\pm$ 0.0358 \\
TNR & Other-Female  & 0.4384 $\pm$ 0.0516 & 0.2741 $\pm$ 0.0276 & 0.3357 $\pm$ 0.0471 \\
TNR & Other-Male  & 0.5071 $\pm$ 0.0490 & 0.3116 $\pm$ 0.0176 & 0.3571 $\pm$ 0.0417 \\
TNR & White-Female  & 0.4164 $\pm$ 0.0294 & 0.1039 $\pm$ 0.0067 & 0.2197 $\pm$ 0.0933 \\
TNR & White-Male  & 0.4694 $\pm$ 0.0281 & 0.1290 $\pm$ 0.0182 & 0.2509 $\pm$ 0.0761 \\
    \addlinespace  \hline  \addlinespace
TPR & Black-Female  & 0.8768 $\pm$ 0.0625 & 0.9658 $\pm$ 0.0145 & 0.9558 $\pm$ 0.0141 \\
TPR & Black-Male  & 0.8147 $\pm$ 0.0493 & 0.9239 $\pm$ 0.0466 & 0.9532 $\pm$ 0.0101 \\
TPR & Other-Female  & 0.9710 $\pm$ 0.0140 & 0.9858 $\pm$ 0.0038 & 0.9851 $\pm$ 0.0052 \\
TPR & Other-Male  & 0.9425 $\pm$ 0.0207 & 0.9805 $\pm$ 0.0062 & 0.9773 $\pm$ 0.0074 \\
TPR & White-Female  & 0.9668 $\pm$ 0.0063 & 0.9976 $\pm$ 0.0008 & 0.9830 $\pm$ 0.0112 \\
TPR & White-Male  & 0.9652 $\pm$ 0.0061 & 0.9965 $\pm$ 0.0013 & 0.9871 $\pm$ 0.0073 \\
    \addlinespace  \hline  \addlinespace
Selection Rate & Black-Female  & 0.6901 $\pm$ 0.0663 & 0.8306 $\pm$ 0.0240 & 0.8137 $\pm$ 0.0162 \\
Selection Rate & Black-Male  & 0.5927 $\pm$ 0.0652 & 0.7785 $\pm$ 0.0432 & 0.7815 $\pm$ 0.0170 \\
Selection Rate & Other-Female  & 0.8605 $\pm$ 0.0224 & 0.9157 $\pm$ 0.0096 & 0.8986 $\pm$ 0.0152 \\
Selection Rate & Other-Male  & 0.8308 $\pm$ 0.0274 & 0.9080 $\pm$ 0.0084 & 0.8942 $\pm$ 0.0150 \\
Selection Rate & White-Female  & 0.8963 $\pm$ 0.0101 & 0.9789 $\pm$ 0.0017 & 0.9457 $\pm$ 0.0261 \\
Selection Rate & White-Male  & 0.8928 $\pm$ 0.0092 & 0.9756 $\pm$ 0.0038 & 0.9475 $\pm$ 0.0186 \\
    \addlinespace  \hline  \addlinespace
ROC AUC & Black-Female  & 0.7364 $\pm$ 0.0150 & 0.6712 $\pm$ 0.0138 & 0.6799 $\pm$ 0.0093 \\
ROC AUC & Black-Male  & 0.7370 $\pm$ 0.0338 & 0.6552 $\pm$ 0.0087 & 0.6834 $\pm$ 0.0191 \\
ROC AUC & Other-Female  & 0.7047 $\pm$ 0.0214 & 0.6300 $\pm$ 0.0125 & 0.6604 $\pm$ 0.0222 \\
ROC AUC & Other-Male  & 0.7248 $\pm$ 0.0149 & 0.6461 $\pm$ 0.0068 & 0.6672 $\pm$ 0.0182 \\
ROC AUC & White-Female  & 0.6916 $\pm$ 0.0122 & 0.5507 $\pm$ 0.0031 & 0.6014 $\pm$ 0.0412 \\
ROC AUC & White-Male  & 0.7173 $\pm$ 0.0119 & 0.5628 $\pm$ 0.0087 & 0.6190 $\pm$ 0.0346 \\
    \bottomrule
\end{tabular}
\caption{LSAC: Mean and standard deviation values for 10 training runs for the best model configurations with and without differential privacy for several metrics regarding all subgroups between race and gender.}\label{tabIntersectionMetricsLSAC}
\end{table*}

\begin{table*}
\centering
\small
\begin{tabular}{cc ccc}
    \toprule
    Metric & Group &  Baseline & Baseline + DP & Best DP Model \\
    \midrule
Accuracy &  Amer-Indian-Eskimo- Female  & 0.9767 $\pm$ 0.0086 & 0.9283 $\pm$ 0.0158 & 0.9667 $\pm$ 0.0000 \\
Accuracy &  Amer-Indian-Eskimo- Male  & 0.8718 $\pm$ 0.0300 & 0.8085 $\pm$ 0.0426 & 0.8803 $\pm$ 0.0120 \\
Accuracy &  Asian-Pac-Islander- Female  & 0.8935 $\pm$ 0.0060 & 0.8290 $\pm$ 0.0591 & 0.8906 $\pm$ 0.0053 \\
Accuracy &  Asian-Pac-Islander- Male  & 0.7849 $\pm$ 0.0100 & 0.7074 $\pm$ 0.0381 & 0.7759 $\pm$ 0.0071 \\
Accuracy &  Black- Female  & 0.9651 $\pm$ 0.0021 & 0.9544 $\pm$ 0.0126 & 0.9620 $\pm$ 0.0019 \\
Accuracy &  Black- Male  & 0.8824 $\pm$ 0.0050 & 0.8115 $\pm$ 0.0286 & 0.8747 $\pm$ 0.0033 \\
Accuracy &  Other- Female  & 0.9738 $\pm$ 0.0075 & 0.9357 $\pm$ 0.0490 & 0.9548 $\pm$ 0.0176 \\
Accuracy &  Other- Male  & 0.8357 $\pm$ 0.0154 & 0.7643 $\pm$ 0.0772 & 0.8229 $\pm$ 0.0100 \\
Accuracy &  White- Female  & 0.9256 $\pm$ 0.0013 & 0.9031 $\pm$ 0.0049 & 0.9245 $\pm$ 0.0008 \\
Accuracy &  White- Male  & 0.8076 $\pm$ 0.0021 & 0.7239 $\pm$ 0.0189 & 0.8058 $\pm$ 0.0012 \\
    \addlinespace  \hline  \addlinespace
Precision &  Amer-Indian-Eskimo- Female  & 0.9400 $\pm$ 0.0966 & 0.5100 $\pm$ 0.3950 & 0.8000 $\pm$ 0.0000 \\
Precision &  Amer-Indian-Eskimo- Male  & 0.6079 $\pm$ 0.2475 & 0.3556 $\pm$ 0.2669 & 0.5952 $\pm$ 0.0471 \\
Precision &  Asian-Pac-Islander- Female  & 0.7273 $\pm$ 0.0817 & 0.3902 $\pm$ 0.1816 & 0.6736 $\pm$ 0.0457 \\
Precision &  Asian-Pac-Islander- Male  & 0.7240 $\pm$ 0.0228 & 0.5604 $\pm$ 0.0477 & 0.7594 $\pm$ 0.0242 \\
Precision &  Black- Female  & 0.6979 $\pm$ 0.0407 & 0.6245 $\pm$ 0.1845 & 0.6120 $\pm$ 0.0234 \\
Precision &  Black- Male  & 0.7373 $\pm$ 0.0217 & 0.5165 $\pm$ 0.0610 & 0.7188 $\pm$ 0.0135 \\
Precision &  Other- Female  & 0.0000 $\pm$ 0.0000 & 0.0250 $\pm$ 0.0791 & 0.0000 $\pm$ 0.0000 \\
Precision &  Other- Male  & 0.6801 $\pm$ 0.0573 & 0.4691 $\pm$ 0.1071 & 0.6381 $\pm$ 0.0426 \\
Precision &  White- Female  & 0.7211 $\pm$ 0.0138 & 0.6234 $\pm$ 0.0504 & 0.7225 $\pm$ 0.0067 \\
Precision &  White- Male  & 0.7183 $\pm$ 0.0132 & 0.5429 $\pm$ 0.0245 & 0.7264 $\pm$ 0.0064 \\
    \addlinespace  \hline  \addlinespace
TNR &  Amer-Indian-Eskimo- Female  & 0.9945 $\pm$ 0.0088 & 0.9836 $\pm$ 0.0200 & 0.9818 $\pm$ 0.0000 \\
TNR &  Amer-Indian-Eskimo- Male  & 0.9738 $\pm$ 0.0207 & 0.8967 $\pm$ 0.0713 & 0.9508 $\pm$ 0.0000 \\
TNR &  Asian-Pac-Islander- Female  & 0.9817 $\pm$ 0.0086 & 0.8933 $\pm$ 0.0884 & 0.9767 $\pm$ 0.0066 \\
TNR &  Asian-Pac-Islander- Male  & 0.8854 $\pm$ 0.0128 & 0.7116 $\pm$ 0.0936 & 0.9197 $\pm$ 0.0122 \\
TNR &  Black- Female  & 0.9895 $\pm$ 0.0023 & 0.9812 $\pm$ 0.0183 & 0.9823 $\pm$ 0.0015 \\
TNR &  Black- Male  & 0.9515 $\pm$ 0.0054 & 0.8635 $\pm$ 0.0601 & 0.9500 $\pm$ 0.0037 \\
TNR &  Other- Female  & 0.9976 $\pm$ 0.0077 & 0.9561 $\pm$ 0.0512 & 0.9780 $\pm$ 0.0180 \\
TNR &  Other- Male  & 0.9625 $\pm$ 0.0056 & 0.8143 $\pm$ 0.1345 & 0.9643 $\pm$ 0.0000 \\
TNR &  White- Female  & 0.9711 $\pm$ 0.0027 & 0.9655 $\pm$ 0.0128 & 0.9723 $\pm$ 0.0011 \\
TNR &  White- Male  & 0.8883 $\pm$ 0.0112 & 0.7139 $\pm$ 0.0450 & 0.8973 $\pm$ 0.0044 \\
    \addlinespace  \hline  \addlinespace
TPR &  Amer-Indian-Eskimo- Female  & 0.7800 $\pm$ 0.0632 & 0.3200 $\pm$ 0.2700 & 0.8000 $\pm$ 0.0000 \\
TPR &  Amer-Indian-Eskimo- Male  & 0.2500 $\pm$ 0.1354 & 0.2700 $\pm$ 0.2058 & 0.4500 $\pm$ 0.0850 \\
TPR &  Asian-Pac-Islander- Female  & 0.3056 $\pm$ 0.0472 & 0.4000 $\pm$ 0.2219 & 0.3167 $\pm$ 0.0527 \\
TPR &  Asian-Pac-Islander- Male  & 0.5881 $\pm$ 0.0156 & 0.6990 $\pm$ 0.1016 & 0.4941 $\pm$ 0.0211 \\
TPR &  Black- Female  & 0.4781 $\pm$ 0.0331 & 0.4188 $\pm$ 0.1744 & 0.5563 $\pm$ 0.0247 \\
TPR &  Black- Male  & 0.5846 $\pm$ 0.0171 & 0.5875 $\pm$ 0.1136 & 0.5500 $\pm$ 0.0166 \\
TPR &  Other- Female  & 0.0000 $\pm$ 0.0000 & 0.1000 $\pm$ 0.3162 & 0.0000 $\pm$ 0.0000 \\
TPR &  Other- Male  & 0.3286 $\pm$ 0.0768 & 0.5643 $\pm$ 0.2368 & 0.2571 $\pm$ 0.0499 \\
TPR &  White- Female  & 0.5754 $\pm$ 0.0175 & 0.4224 $\pm$ 0.0823 & 0.5565 $\pm$ 0.0069 \\
TPR &  White- Male  & 0.6289 $\pm$ 0.0228 & 0.7461 $\pm$ 0.0459 & 0.6032 $\pm$ 0.0071 \\
    \bottomrule
\end{tabular}
\caption{Adult: Mean and standard deviation values for 10 training runs for the best model configurations with and without differential privacy for Accuracy, Precision True Negative Rate, and True Positive Rate on all subgroups between race and gender.}\label{tabIntersectionMetricsAdult}
\end{table*}

\begin{table*}
\centering
\small
\begin{tabular}{cc ccc}
    \toprule
    Metric & Group &  Baseline & Baseline + DP & Best DP Model \\
    \midrule
Selection Rate &  Amer-Indian-Eskimo- Female  & 0.0700 $\pm$ 0.0105 & 0.0417 $\pm$ 0.0379 & 0.0833 $\pm$ 0.0000 \\
Selection Rate &  Amer-Indian-Eskimo- Male  & 0.0577 $\pm$ 0.0215 & 0.1268 $\pm$ 0.0858 & 0.1056 $\pm$ 0.0120 \\
Selection Rate &  Asian-Pac-Islander- Female  & 0.0558 $\pm$ 0.0123 & 0.1449 $\pm$ 0.1001 & 0.0616 $\pm$ 0.0115 \\
Selection Rate &  Asian-Pac-Islander- Male  & 0.2746 $\pm$ 0.0100 & 0.4271 $\pm$ 0.0927 & 0.2201 $\pm$ 0.0135 \\
Selection Rate &  Black- Female  & 0.0328 $\pm$ 0.0032 & 0.0379 $\pm$ 0.0243 & 0.0434 $\pm$ 0.0018 \\
Selection Rate &  Black- Male  & 0.1494 $\pm$ 0.0059 & 0.2215 $\pm$ 0.0697 & 0.1442 $\pm$ 0.0051 \\
Selection Rate &  Other- Female  & 0.0024 $\pm$ 0.0075 & 0.0452 $\pm$ 0.0520 & 0.0214 $\pm$ 0.0176 \\
Selection Rate &  Other- Male  & 0.0957 $\pm$ 0.0166 & 0.2614 $\pm$ 0.1472 & 0.0800 $\pm$ 0.0100 \\
Selection Rate &  White- Female  & 0.0917 $\pm$ 0.0042 & 0.0791 $\pm$ 0.0203 & 0.0885 $\pm$ 0.0015 \\
Selection Rate &  White- Male  & 0.2726 $\pm$ 0.0146 & 0.4292 $\pm$ 0.0444 & 0.2584 $\pm$ 0.0052 \\
    \addlinespace  \hline  \addlinespace
ROC AUC &  Amer-Indian-Eskimo- Female  & 0.8873 $\pm$ 0.0310 & 0.6518 $\pm$ 0.1280 & 0.8909 $\pm$ 0.0000 \\
ROC AUC &  Amer-Indian-Eskimo- Male  & 0.6119 $\pm$ 0.0717 & 0.5834 $\pm$ 0.0783 & 0.7004 $\pm$ 0.0425 \\
ROC AUC &  Asian-Pac-Islander- Female  & 0.6436 $\pm$ 0.0211 & 0.6467 $\pm$ 0.0841 & 0.6467 $\pm$ 0.0243 \\
ROC AUC &  Asian-Pac-Islander- Male  & 0.7367 $\pm$ 0.0101 & 0.7053 $\pm$ 0.0280 & 0.7069 $\pm$ 0.0086 \\
ROC AUC &  Black- Female  & 0.7338 $\pm$ 0.0161 & 0.7000 $\pm$ 0.0807 & 0.7693 $\pm$ 0.0124 \\
ROC AUC &  Black- Male  & 0.7680 $\pm$ 0.0085 & 0.7255 $\pm$ 0.0287 & 0.7500 $\pm$ 0.0077 \\
ROC AUC &  Other- Female  & 0.4988 $\pm$ 0.0039 & 0.5280 $\pm$ 0.1550 & 0.4890 $\pm$ 0.0090 \\
ROC AUC &  Other- Male  & 0.6455 $\pm$ 0.0381 & 0.6893 $\pm$ 0.0791 & 0.6107 $\pm$ 0.0250 \\
ROC AUC &  White- Female  & 0.7732 $\pm$ 0.0077 & 0.6939 $\pm$ 0.0355 & 0.7644 $\pm$ 0.0032 \\
ROC AUC &  White- Male  & 0.7586 $\pm$ 0.0062 & 0.7300 $\pm$ 0.0094 & 0.7503 $\pm$ 0.0016 \\
    \bottomrule
\end{tabular}
\caption{Adult: Mean and standard deviation values for 10 training runs for the best model configurations with and without differential privacy for Selection Rate and ROC AUC on all subgroups between race and gender.}\label{tabIntersectionMetricsAdult2}
\end{table*}

\begin{table*}
\centering
\small
\begin{tabular}{cc ccc}
    \toprule
    Metric & Group &  Baseline & Baseline + DP & Best DP Model \\
    \midrule

%%%%%%%%%%%%%%%%%%%%%%%%%%%%%%
Accuracy & Black-Female  & 0.6628 $\pm$ 0.0122 & 0.5387 $\pm$ 0.0533 & 0.6558 $\pm$ 0.0216 \\
Accuracy & Black-Male  & 0.6245 $\pm$ 0.0092 & 0.5172 $\pm$ 0.0387 & 0.6103 $\pm$ 0.0071 \\
Accuracy & Other-Female  & 0.7178 $\pm$ 0.0150 & 0.6289 $\pm$ 0.0777 & 0.7000 $\pm$ 0.0189 \\
Accuracy & Other-Male  & 0.6265 $\pm$ 0.0157 & 0.5253 $\pm$ 0.0422 & 0.6378 $\pm$ 0.0105 \\
Accuracy & White-Female  & 0.6847 $\pm$ 0.0203 & 0.5682 $\pm$ 0.0444 & 0.6506 $\pm$ 0.0180 \\
Accuracy & White-Male  & 0.6289 $\pm$ 0.0072 & 0.5051 $\pm$ 0.0404 & 0.6266 $\pm$ 0.0151 \\
    \addlinespace  \hline  \addlinespace
Precision & Black-Female  & 0.5803 $\pm$ 0.0216 & 0.4145 $\pm$ 0.0444 & 0.5578 $\pm$ 0.0305 \\
Precision & Black-Male  & 0.6640 $\pm$ 0.0065 & 0.5862 $\pm$ 0.0369 & 0.6705 $\pm$ 0.0060 \\
Precision & Other-Female  & 0.3300 $\pm$ 0.0740 & 0.2969 $\pm$ 0.1149 & 0.3537 $\pm$ 0.0489 \\
Precision & Other-Male  & 0.5953 $\pm$ 0.0289 & 0.4523 $\pm$ 0.0452 & 0.6005 $\pm$ 0.0216 \\
Precision & White-Female  & 0.6178 $\pm$ 0.0422 & 0.4290 $\pm$ 0.0534 & 0.5488 $\pm$ 0.0353 \\
Precision & White-Male  & 0.5893 $\pm$ 0.0094 & 0.4513 $\pm$ 0.0365 & 0.5870 $\pm$ 0.0195 \\
    \addlinespace  \hline  \addlinespace
TNR & Black-Female  & 0.7861 $\pm$ 0.0202 & 0.6041 $\pm$ 0.1355 & 0.7303 $\pm$ 0.0277 \\
TNR & Black-Male  & 0.5013 $\pm$ 0.0111 & 0.4361 $\pm$ 0.1050 & 0.5619 $\pm$ 0.0113 \\
TNR & Other-Female  & 0.9000 $\pm$ 0.0152 & 0.7147 $\pm$ 0.1127 & 0.8382 $\pm$ 0.0208 \\
TNR & Other-Male  & 0.7887 $\pm$ 0.0223 & 0.5739 $\pm$ 0.1063 & 0.7613 $\pm$ 0.0292 \\
TNR & White-Female  & 0.8436 $\pm$ 0.0226 & 0.6536 $\pm$ 0.0766 & 0.8000 $\pm$ 0.0329 \\
TNR & White-Male  & 0.7184 $\pm$ 0.0100 & 0.5028 $\pm$ 0.1645 & 0.7193 $\pm$ 0.0160 \\
    \addlinespace  \hline  \addlinespace
TPR & Black-Female  & 0.4675 $\pm$ 0.0237 & 0.4351 $\pm$ 0.1092 & 0.5377 $\pm$ 0.0336 \\
TPR & Black-Male  & 0.7137 $\pm$ 0.0139 & 0.5759 $\pm$ 0.0883 & 0.6454 $\pm$ 0.0114 \\
TPR & Other-Female  & 0.1545 $\pm$ 0.0439 & 0.3636 $\pm$ 0.1660 & 0.2727 $\pm$ 0.0429 \\
TPR & Other-Male  & 0.4112 $\pm$ 0.0159 & 0.4607 $\pm$ 0.0999 & 0.4738 $\pm$ 0.0202 \\
TPR & White-Female  & 0.4197 $\pm$ 0.0328 & 0.4258 $\pm$ 0.0595 & 0.4015 $\pm$ 0.0229 \\
TPR & White-Male  & 0.5149 $\pm$ 0.0145 & 0.5081 $\pm$ 0.1352 & 0.5085 $\pm$ 0.0247 \\
    \addlinespace  \hline  \addlinespace
Selection Rate & Black-Female  & 0.3121 $\pm$ 0.0181 & 0.4111 $\pm$ 0.1206 & 0.3734 $\pm$ 0.0212 \\
Selection Rate & Black-Male  & 0.6234 $\pm$ 0.0095 & 0.5709 $\pm$ 0.0874 & 0.5583 $\pm$ 0.0090 \\
Selection Rate & Other-Female  & 0.1133 $\pm$ 0.0164 & 0.3044 $\pm$ 0.1084 & 0.1889 $\pm$ 0.0189 \\
Selection Rate & Other-Male  & 0.2972 $\pm$ 0.0130 & 0.4410 $\pm$ 0.0962 & 0.3398 $\pm$ 0.0244 \\
Selection Rate & White-Female  & 0.2551 $\pm$ 0.0170 & 0.3761 $\pm$ 0.0601 & 0.2756 $\pm$ 0.0259 \\
Selection Rate & White-Male  & 0.3842 $\pm$ 0.0096 & 0.5020 $\pm$ 0.1497 & 0.3809 $\pm$ 0.0130 \\
    \addlinespace  \hline  \addlinespace
ROC AUC & Black-Female  & 0.6268 $\pm$ 0.0122 & 0.5196 $\pm$ 0.0376 & 0.6340 $\pm$ 0.0220 \\
ROC AUC & Black-Male  & 0.6075 $\pm$ 0.0087 & 0.5060 $\pm$ 0.0395 & 0.6036 $\pm$ 0.0070 \\
ROC AUC & Other-Female  & 0.5273 $\pm$ 0.0226 & 0.5392 $\pm$ 0.0787 & 0.5555 $\pm$ 0.0238 \\
ROC AUC & Other-Male  & 0.6000 $\pm$ 0.0151 & 0.5173 $\pm$ 0.0388 & 0.6175 $\pm$ 0.0082 \\
ROC AUC & White-Female  & 0.6317 $\pm$ 0.0215 & 0.5397 $\pm$ 0.0385 & 0.6008 $\pm$ 0.0148 \\
ROC AUC & White-Male  & 0.6166 $\pm$ 0.0075 & 0.5055 $\pm$ 0.0280 & 0.6139 $\pm$ 0.0157 \\
    \bottomrule
\end{tabular}
\caption{Compas: Mean and standard deviation values for 10 training runs for the best model configurations with and without differential privacy for several metrics regarding all subgroups between race and gender.}\label{tabIntersectionMetricsCompas}
\end{table*}

\end{document}